\documentclass{article}

\usepackage{microtype}
\usepackage{graphicx}
\usepackage{subfigure}
\usepackage{booktabs} % for professional tables

\usepackage{hyperref}

\usepackage[accepted]{icml2021}

\usepackage{nicefrac}
\usepackage{color}
\usepackage{multirow}
\usepackage{makecell}
\usepackage{colortbl}
\usepackage{amssymb}
\usepackage{amsmath}
\usepackage{mathrsfs}
\usepackage{bm}
\usepackage{tikz}
\usepackage{xspace}
\xspaceaddexceptions{[]\{\}}
\usepackage{tablefootnote}
\usepackage{setspace}

\newcommand{\cA}{{\mathcal A}}

\newcommand{\cD}{{\mathcal D}}

\newtheorem{theorem}{Theorem}
\newtheorem{lemma}{Lemma}
\newtheorem{corollary}{Corollary}

\newtheorem{definition}{Definition}
\newtheorem{assumption}{Assumption}

{\hspace*{\fill}$\Box$\par}
{\hspace*{\fill}$\Box$\par\vspace{4mm}}
\newenvironment{proofof}[1]{\smallskip\noindent{\bf \em Proof of #1.}}%
{\hspace*{\fill}$\Box$\par}

\renewcommand{\algorithmicrequire}{ \textbf{Input:}} % Use Input in the format of Algorithm
\renewcommand{\algorithmicensure}{ \textbf{Output:}} % Use Output in the format of Algorithm

\newcommand{\eat}[1]{}
\definecolor{bgcolor}{rgb}{0.66,0.88,1.00}
\definecolor{bgcolor2}{rgb}{0.66,0.88,0.50}

\newcommand{\E}{{\mathbb{E}}}
\newcommand{\R}{{\mathbb{R}}}
\newcommand{\inner}[2]{\langle #1,#2 \rangle}

\newcommand{\ns}[1]{\| #1 \|^2}
\newcommand{\nsB}[1]{\left\| #1 \right\|^2}

\newcommand{\n}[1]{\| #1 \|}

\newcommand{\hx}{\widehat{x}}

\newcommand{\alglabel}{%
	\addtocounter{ALC@line}{-1}% Reduce line counter by 1
	\refstepcounter{ALC@line}% Increment line counter with reference capability
	\label% Regular \label
}

\usepackage{todonotes}

\newcommand{\page}{{\sf PAGE}\xspace}
\newcommand{\compactify}{}

\newcommand{\fgap}{{\Delta_0}}

\newcommand{\xt}{x^t}
\newcommand{\xtn}{x^{t+1}}
\newcommand{\bxtn}{\bar{x}^{t+1}}
\newcommand{\gt}{g^t}
\newcommand{\gtn}{g^{t+1}}
\newcommand{\pt}{\Phi_t}
\newcommand{\ptn}{\Phi_{t+1}}
\newcommand{\EB}[1]{\E\left[ #1 \right]}
\newcommand{\newl}{\notag \\ &\qquad }
\newcommand{\newll}{\notag \\ &\qquad \qquad }
\icmltitlerunning{PAGE: A Simple and Optimal Probabilistic Gradient Estimator for Nonconvex Optimization}

\begin{document}

\twocolumn[
\icmltitle{PAGE: A Simple and Optimal Probabilistic Gradient Estimator for \\ Nonconvex Optimization}

% It is OKAY to include author information, even for blind
% submissions: the style file will automatically remove it for you
% unless you've provided the [accepted] option to the icml2021
% package.

% List of affiliations: The first argument should be a (short)
% identifier you will use later to specify author affiliations
% Academic affiliations should list Department, University, City, Region, Country
% Industry affiliations should list Company, City, Region, Country

% You can specify symbols, otherwise they are numbered in order.
% Ideally, you should not use this facility. Affiliations will be numbered
% in order of appearance and this is the preferred way.
%\icmlsetsymbol{equal}{*}

\begin{icmlauthorlist}
\icmlauthor{Zhize Li}{kaust}
\icmlauthor{Hongyan Bao}{kaust}
\icmlauthor{Xiangliang Zhang}{kaust}
\icmlauthor{Peter Richt\'{a}rik}{kaust}
\end{icmlauthorlist}

\icmlaffiliation{kaust}{King Abdullah University of Science and Technology, Thuwal, Kingdom of Saudi Arabia}

\icmlcorrespondingauthor{Zhize Li}{zhize.li@kaust.edu.sa}

% You may provide any keywords that you
% find helpful for describing your paper; these are used to populate
% the "keywords" metadata in the PDF but will not be shown in the document
\icmlkeywords{nonconvex optimization, optimal method, lower bound, finite-sum optimization, online optimization}

\vskip 0.3in
]

% this must go after the closing bracket ] following \twocolumn[ ...

% This command actually creates the footnote in the first column
% listing the affiliations and the copyright notice.
% The command takes one argument, which is text to display at the start of the footnote.
% The \icmlEqualContribution command is standard text for equal contribution.
% Remove it (just {}) if you do not need this facility.

\printAffiliationsAndNotice{}  % leave blank if no need to mention equal contribution
%\printAffiliationsAndNotice{\icmlEqualContribution} % otherwise use the standard text.

\begin{abstract}
 In this paper, we propose a novel stochastic gradient estimator---ProbAbilistic Gradient Estimator (\page)---for nonconvex optimization. 
 \page is easy to implement as it is designed via a small adjustment to vanilla SGD:  in each iteration, \page uses the vanilla minibatch SGD update with probability $p_t$ or reuses the previous gradient with a small adjustment, at a much lower computational cost, with probability $1-p_t$. We give a simple formula for the optimal choice of $p_t$. Moreover, we prove the first tight lower bound $\Omega(n+\frac{\sqrt{n}}{\epsilon^2})$ for nonconvex finite-sum problems, which also leads to a tight lower bound $\Omega(b+\frac{\sqrt{b}}{\epsilon^2})$ for nonconvex online problems, where $b:= \min\{\frac{\sigma^2}{\epsilon^2}, n\}$.
 Then, we show that \page obtains the optimal convergence results $O(n+\frac{\sqrt{n}}{\epsilon^2})$ (finite-sum) and $O(b+\frac{\sqrt{b}}{\epsilon^2})$ (online) matching our lower bounds for both nonconvex finite-sum and online problems.
  Besides, we also show that for nonconvex functions satisfying the Polyak-\L{}ojasiewicz (PL) condition, \page can automatically switch to a faster linear convergence rate $O(\cdot\log \frac{1}{\epsilon})$. 
  Finally, we conduct several deep learning experiments (e.g., LeNet, VGG, ResNet) on real datasets in PyTorch showing that \page not only converges much faster than SGD in training but also achieves the higher test accuracy, validating the optimal theoretical results and confirming the practical superiority of \page.
\end{abstract}

\section{Introduction}
Nonconvex optimization is ubiquitous across many domains of machine learning, including robust regression, low rank matrix recovery, sparse recovery and supervised learning \citep{NonconvexBook}. Driven by the applied success of deep neural networks \citep{DL-Nature-2015}, and the critical place nonconvex optimization plays in training them,  research in nonconvex optimization has been undergoing a renaissance~\citep{ghadimi2013stochastic, ghadimi2016mini, zhou2018stochastic, fang2018spider, zhize2019ssrgd, li2020unified}.  

\subsection{The problem} 
\label{sec:problem}
Motivated by this development,  we consider the general optimization problem
\begin{equation}\label{eq:prob}
\min_{x\in \R^d}   f(x),
\end{equation}
where $f:
\R^d\to \R$ is a differentiable and possibly nonconvex  function. We are interested in functions  having the \emph{finite-sum} form
\begin{align}
f(x) := \frac{1}{n}\sum_{i=1}^n{f_i(x)}, \label{prob:finite}
\end{align}
where the functions $f_i$  are also differentiable and possibly nonconvex.
Form \eqref{prob:finite} captures the standard empirical risk minimization  problems in machine learning \citep{shai_book}. Moreover, if the number of data samples $n$ is very large or even infinite, e.g., in the online/streaming case, then $f(x)$ usually is modeled via  the \emph{online} form
\begin{align}
f(x) := \E_{\zeta\sim \cD}[F(x,\zeta)],  \label{prob:online}
\end{align}
which we also consider in this work. For notational convenience, we adopt the notation of the finite-sum form \eqref{prob:finite} in the descriptions and algorithms in the rest of this paper. However,  our results apply to the  online form \eqref{prob:online} as well by letting $f_i(x) := F(x, \zeta_i)$ and treating $n$ as a very large value or even infinite.

%In this paper, we focus on finding an $\epsilon$-approximate solution which is defined as follows:
%\begin{definition}
%	A point $x$ is called an $\epsilon$-approximate solution for problem \eqref{eq:prob} if 
%	$\n{\nabla f(x)} \leq \epsilon$.
%\end{definition}

\subsection{Gradient complexity}

To measure the efficiency of algorithms for solving the nonconvex optimization  problem \eqref{eq:prob}, it is standard to bound the number of stochastic gradient computations needed to find a solution of suitable characteristics. In this paper we use the standard term \emph{gradient complexity} to describe such bounds. In particular, our goal will be to find a (possibly random) point $\hx \in \R^d$ such that $\E[{\n{\nabla f(\hx)}}] \leq \epsilon$, where the expectation is with respect to the randomness inherent in the algorithm. We use the term  {\em $\epsilon$-approximate solution} to refer to such a point $\hx$.

Two of the most classical gradient complexity results  for solving problem \eqref{eq:prob} are those for  gradient descent (GD) and stochastic gradient descent (SGD).  In particular,  the gradient complexity of GD is $O(\nicefrac{n}{\epsilon^2})$ in this nonconvex regime, and  assuming that the stochastic gradient satisfies a (uniform) bounded variance  assumption (Assumption~\ref{asp:bv}), the gradient complexity of SGD is $O(\nicefrac{1}{\epsilon^4})$.  Note that although SGD has a worse dependence on $\epsilon$,  it typically only needs to compute a constant minibatch of stochastic gradients in each iteration instead of the full batch (i.e., $n$ stochastic gradients) used in GD. Hence, SGD is better than GD if the number of data samples $n$ is very large or the error tolerance $\epsilon$ is not very small.

There has been extensive research in designing gradient-type methods with an improved dependence on $n$ and/or $\epsilon$ \citep{nesterov2014introductory, nemirovski2009robust, ghadimi2013stochastic, ghadimi2016mini}. In particular, the  SVRG method of \citet{johnson2013accelerating},  the SAGA method of  \citet{defazio2014saga} and the SARAH  method of \citet{nguyen2017sarah} are representatives of what is by now a large class of {\em variance-reduced} methods, which have played a particularly important role in this effort.  However, the analyses in these papers focused on the convex regime. Furthermore, several accelerated (momentum) methods have been designed as well \citep{nesterov83, lan2015optimal, lin2015universal, lan2018random, allen2017katyusha, zhize2020anderson, zhize2019unified, li2020acceleration, li2021anita}, with or without variance reduction. There are also some lower bounds given by \citep{lan2015optimal, woodworth2016tight, xie2019general}.

Coming back to problem \eqref{eq:prob} in the nonconvex regime studied in this paper, interesting recent development starts with the work of \citet{reddi2016stochastic}, and \citet{allen2016variance}, who have concurrently shown that if $f$ has the  finite-sum form~\eqref{prob:finite},  a suitably designed minibatch version of SVRG enjoys the gradient complexity $O(n+\nicefrac{n^{2/3}}{\epsilon^2})$, which is an improvement on the $O(\nicefrac{n}{\epsilon^2})$ gradient complexity of GD. Subsequently, other variants of SVRG were shown to posses the same improved rate, including those developed by \citep{lei2017non, li2018simple, ge2019stable, nonconvex_arbitrary, qian2019svrg}. 
More recently, \citet{fang2018spider} proposed the  SPIDER method, and \citet{zhou2018stochastic}  proposed the SNVRG method, both of them  further improve the gradient complexity to $O(n+ \nicefrac{\sqrt{n}}{\epsilon^2})$. Further variants of the SARAH method (e.g., \citealp{wang2018spiderboost, zhize2019ssrgd, pham2019proxsarah, li2020convergence, horvath2020adaptivity, li2021zerosarah}) which also achieve the same $O(n+ \nicefrac{\sqrt{n}}{\epsilon^2})$ gradient complexity have been developed. 
Also there are some lower bounds given by \citep{fang2018spider, zhou2019lower, arjevani2019lower}. See Table~\ref{table:nonconvex} for an overview of results.

\section{Our Contributions}

As we show in through this work, despite enormous effort by the community to design efficient methods for solving \eqref{eq:prob} in the nonconvex regime, there is still a considerable gap in our understanding. First, while optimal methods for \eqref{eq:prob} in the finite-sum regime exist (e.g., SPIDER \citep{fang2018spider}, SpiderBoost \citep{wang2018spiderboost}, SARAH \citep{pham2019proxsarah}, SSRGD \citep{zhize2019ssrgd}), the known lower bound $\Omega(\nicefrac{\sqrt{n}}{\epsilon^2})$ \citep{fang2018spider} used to establish their optimality works only for $n\leq O(\nicefrac{1}{\epsilon^4})$, i.e.,  in the small data regime (see Table~\ref{table:nonconvex}). Moreover, these methods are unnecessarily complicated, often with a double loop structure, and reliance on several hyperparameters. Besides, there is also no tight lower bound to show the optimality of optimal methods in the online regime.

\begin{table*}[t]
	\vspace{-3mm}
	\begin{minipage}{\textwidth}
		\centering
		\footnotesize
		\caption{Stochastic gradient complexity for finding an $\epsilon$-approximate solution $\E[\n{\nabla f(\hx)}] \leq \epsilon$ for nonconvex problems} 
		\label{table:nonconvex}
		\vspace{1mm}
		\renewcommand{\arraystretch}{1.6}
		\begin{tabular}{|c|c|c|c|}
			\hline
			\bf Problem
			& \bf Assumption
			& \bf Algorithm or Lower Bound
			& \bf Gradient complexity\\
			\hline
			
			%%%
			%%% finite-sum
			%%%
			Finite-sum \eqref{prob:finite} 
			& Asp. \ref{asp:smooth}
			& GD \citep{nesterov2014introductory} 
			& $O(\frac{n}{\epsilon^2})$ \\
			\hline
			
			Finite-sum \eqref{prob:finite} 
			& Asp. \ref{asp:smooth}
			& \makecell{SVRG \citep{allen2016variance, reddi2016stochastic} \\
				SCSG \citep{lei2017non}, \\
				SVRG+ \citep{li2018simple}}
			& $O(n+ \frac{n^{2/3}}{\epsilon^2})$ \\
			\hline
			
			Finite-sum \eqref{prob:finite} 
			& Asp. \ref{asp:smooth}
			&  \makecell{SNVRG \citep{zhou2018stochastic}, \\
				Geom-SARAH \citep{horvath2020adaptivity}}
			& $\widetilde{O}\left(n+ \frac{\sqrt{n}}{\epsilon^2}\right)$ \\
			\hline
			
			Finite-sum \eqref{prob:finite} 
			& Asp. \ref{asp:smooth}
			&  \makecell{ SPIDER \citep{fang2018spider}, \\
				SpiderBoost \citep{wang2018spiderboost},\\
				SARAH \citep{pham2019proxsarah},  \\
				SSRGD \citep{zhize2019ssrgd}}
			& $O(n+ \frac{\sqrt{n}}{\epsilon^2})$ \\
			\hline
			
			\rowcolor{bgcolor}
			Finite-sum \eqref{prob:finite} 
			& Asp. \ref{asp:smooth}
			& \page (this paper) 
			& $O(n+\frac{\sqrt{n}}{\epsilon^2})$ \\
			\hline
			
			Finite-sum \eqref{prob:finite} 
			& Asp. \ref{asp:smooth}
			& Lower bound \citep{fang2018spider}
			& $\Omega(\frac{\sqrt{n}}{\epsilon^2})$~  if $n\leq O(\frac{1}{\epsilon^4})$ \\
			\hline
			
			\rowcolor{bgcolor}
			Finite-sum \eqref{prob:finite} 
			& Asp. \ref{asp:smooth}
			& Lower bound (this paper) 
			& $\Omega(n+\frac{\sqrt{n}}{\epsilon^2})$ \\
			\hline
			
			\rowcolor{bgcolor2}
			Finite-sum \eqref{prob:finite} 
			& Asp. \ref{asp:smooth} and \ref{asp:pl} (PL setting)
			& \page (this paper) 
			& $O\left((n+\sqrt{n}\kappa)\log\frac{1}{\epsilon}\right)$
			\footnote{Note that \page can switch to a faster \emph{linear convergence} $O(\cdot\log \frac{1}{\epsilon})$ instead of sublinear rate $O(\cdot\frac{1}{\epsilon^2})$ by exploiting the local structure of the objective function via the PL condition (Assumption \ref{asp:pl}).} \\
			\hline
			\hline

			%%%
			%%% online
			%%%
			Online \eqref{prob:online}~\footnote{Note that we refer the online problem \eqref{prob:online} as the finite-sum problem \eqref{prob:finite} with large or infinite $n$ as discussed in the introduction Section \ref{sec:problem}. In this online case, the full gradient may not be available (e.g., if $n$ is infinite), thus the bounded variance of stochastic gradient Assumption \ref{asp:bv} is needed in this case.}
			& Asp. \ref{asp:bv} and \ref{asp:smooth}
			& \makecell{SGD (\citealp{ghadimi2016mini};\\ 
				\citealp{khaled2020better, li2020unified})}
			& $O(\frac{\sigma^2}{\epsilon^4})$ \\
			\hline
			
			Online \eqref{prob:online} 
			& Asp. \ref{asp:bv} and \ref{asp:smooth}
			& \makecell{ SCSG \citep{lei2017non}, \\
				SVRG+ \citep{li2018simple}}
			& $O(b+\frac{b^{2/3}}{\epsilon^2})$ \\
			\hline
			
			Online \eqref{prob:online} 
			& Asp. \ref{asp:bv} and \ref{asp:smooth}
			& \makecell{SNVRG \citep{zhou2018stochastic}, \\
				Geom-SARAH \citep{horvath2020adaptivity}}
			& $\widetilde{O}\left(b+ \frac{\sqrt{b}}{\epsilon^2}\right)$ \\
			\hline
			
			Online \eqref{prob:online} 
			& Asp. \ref{asp:bv} and \ref{asp:smooth}
			&  \makecell{ SPIDER \citep{fang2018spider}, \\
				SpiderBoost \citep{wang2018spiderboost},\\
				SARAH \citep{pham2019proxsarah}, \\
				SSRGD \citep{zhize2019ssrgd}}
			& $O(b+ \frac{\sqrt{b}}{\epsilon^2})$ \\
			\hline
			
			\rowcolor{bgcolor}
			Online \eqref{prob:online} 
			& Asp. \ref{asp:bv} and \ref{asp:smooth}
			& \page (this paper) 
			& $O(b+\frac{\sqrt{b}}{\epsilon^2})$ 
			\footnote{In the online case, $b:= \min\{\frac{\sigma^2}{\epsilon^2}, n\}$, and $\sigma$ is defined in Assumption \ref{asp:bv}. 
				If $n$ is very large, i.e., $b:=\min\{\frac{\sigma^2}{\epsilon^2}, n\}=\frac{\sigma^2}{\epsilon^2}$, then $O(b+\frac{\sqrt{b}}{\epsilon^2})=O(\frac{\sigma^2}{\epsilon^2} + \frac{\sigma}{\epsilon^3})$ is better than the rate $O(\frac{\sigma^2}{\epsilon^4})$ of SGD by a factor of $\frac{1}{\epsilon^2}$ or $\frac{\sigma}{\epsilon}$.}\\
			\hline
			
			\rowcolor{bgcolor}
			Online \eqref{prob:online} 
			& Asp. \ref{asp:bv} and \ref{asp:smooth}
			& Lower bound (this paper) 
			& $\Omega(b+\frac{\sqrt{b}}{\epsilon^2})$ \\
			\hline
			
			\rowcolor{bgcolor2}
			Online \eqref{prob:online} 
			& Asp. \ref{asp:bv}, \ref{asp:smooth} and \ref{asp:pl} (PL setting)
			& \page (this paper) 
			& $O\left((b+\sqrt{b}\kappa)\log\frac{1}{\epsilon}\right)$\\
			\hline
		\end{tabular}
		
	\end{minipage}
	\vspace{-2.5mm}
\end{table*}

In this paper, we resolve the above issues by designing a simple ProbAbilistic Gradient Estimator (\page) described in Algorithm \ref{alg:page} for achieving optimal convergence results in nonconvex optimization.  Moreover, \page is very simple and easy to implement. In each iteration, \page uses minibatch SGD update with probability $p_t$, or reuses the previous gradient with a small adjustment (at a low computational cost) with probability $1-p_t$ (see Line \ref{line:grad} of Algorithm~\ref{alg:page}).
We would like to highlight the following results:

$\bullet$We provide tight lower bounds to close the gap for both nonconvex finite-sum problem \eqref{prob:finite} and online problem \eqref{prob:online} (see Theorem~\ref{thm:lb} and Corollary~\ref{cor:lbonline}). 
Our lower bounds are based and inspired by recent work \citep{fang2018spider, arjevani2019lower}. 
Then we show the optimality of \page by proving that \page achieves the optimal convergence results matching our lower bounds for both nonconvex finite-sum problem \eqref{prob:finite} and online problem \eqref{prob:online} (see Corollaries~\ref{cor:finite} and \ref{cor:online}).  
See Table \ref{table:nonconvex} for a detailed comparison.

$\bullet$ Moreover, we show that \page can automatically switch to a faster \emph{linear convergence} $O(\cdot\log \frac{1}{\epsilon})$ by exploiting the local structure of the objective function, via the PL condition (Assumption \ref{asp:pl}), although the objective function $f$ is globally nonconvex. See the middle and the last row of Table \ref{table:nonconvex} (highlighted with green color).
For example,  \page automatically switches from the  sublinear rate $O(n+\nicefrac{\sqrt{n}}{\epsilon^2})$ to the faster linear rate $O((n+\sqrt{n}\kappa)\log\frac{1}{\epsilon})$ for nonconvex finite-sum problem \eqref{prob:finite}. % (see the Remark after Corollary \ref{cor:finite-pl}).  

$\bullet$  \page is simple and easy to implement via a small adjustment to vanilla minibatch SGD, and takes a lower computational cost than SGD (i.e., $p_t=1$ in Algorithm \ref{alg:page}) since $b'<b$. 	We conduct several deep learning experiments (e.g., LeNet, VGG, ResNet) on real datasets in PyTorch showing that \page indeed not only converges much faster than SGD in training but also achieves higher test accuracy. This validates our theoretical results and confirms the practical superiority of \page.

\begin{algorithm*}[t]
	\caption{ProbAbilistic Gradient Estimator (\page)}
	\label{alg:page}
	\begin{algorithmic}[1]
		\REQUIRE ~%Input
		initial point $x^0$, stepsize $\eta$, minibatch size $b$,~ $b'<b$, probability $\{p_t\} \in (0,1]$
		\STATE $g^{0} = \frac{1}{b} \sum_{i\in I} \nabla f_i(x^0)$  \quad {\footnotesize ~//~$I$ denotes random minibatch samples with $|I|=b$} \alglabel{line:sgd}
		\FOR {$t=0,1,2,\ldots$}
		\STATE $x^{t+1} = x^t - \eta g^t$  \alglabel{line:update}
		\STATE $g^{t+1} = \begin{cases}
		\frac{1}{b} \sum \limits_{i\in I} \nabla f_i(x^{t+1}) &\text{with probability } p_t\\
		g^{t}+\frac{1}{b'} \sum \limits_{i\in I'} (\nabla f_i(x^{t+1})- \nabla f_i(x^{t})) &\text{with probability } 1-p_t
		\end{cases}$   \alglabel{line:grad}
		\ENDFOR
		\ENSURE $\hx_T$ chosen uniformly from $\{x^t\}_{t\in [T]}$  
	\end{algorithmic}
\end{algorithm*}

\subsection{The PAGE gradient estimator}

In this section, we describe \page, an SGD variant employing a new, simple and optimal gradient estimator (see Algorithm \ref{alg:page}).
In particular, \page was inspired by algorithmic design elements coming from methods such as SARAH \citep{nguyen2017sarah}, SPIDER \citep{fang2018spider}, SSRGD \citep{zhize2019ssrgd} (usage of a  recursive estimator), and  L-SVRG \citep{kovalev2019don} and SAGD \citep{bibi2018improving} (probabilistic switching between two estimators to avoid a double loop structure).

In each iteration, the gradient estimator $g^{t+1}$ of \page is defined in  Line \ref{line:grad} of Algorithm \ref{alg:page}, which indicates that
%\begin{align} \label{eq:grad}
%g^{t+1} = \begin{cases}
%\frac{1}{b} \sum_{i\in I} \nabla f_i(x^{t+1}) &\text{with probability } p,\\
%g^{t}+\frac{1}{b'} \sum_{i\in I'} (\nabla f_i(x^{t+1})- \nabla f_i(x^{t})) &\text{with probability } 1-p.
%\end{cases}
%\end{align}
\page uses the vanilla minibatch SGD update with probability $p_t$, and reuses the previous gradient $g^{t}$ with a small adjustment (which  lowers the computational cost since $b' \ll b$) with probability $1-p_t$. In particular, the  $p_t \equiv 1$ case reduces to vanilla minibatch SGD, and to GD if we further set the minibatch size to $b=n$.
We give a simple formula for the optimal choice of $p_t$, i.e., $p_t\equiv \frac{b'}{b+b'}$ is enough for \page to obtain the optimal convergence rates.
More details can be found in the convergence results of Section \ref{sec:result}.

Note that \page with constant probability $p_t\equiv p$ can be reduced to an equivalent form of the double loop algorithm with geometric distribution Geom-SARAH \citep{horvath2020adaptivity}, but our single-loop \page is more flexible and also leads to simpler and better analysis.
Similar to L-SVRG \citep{kovalev2019don} which switches between GD and SVRG probabilistically, L2S \citep{li2020convergence} switches between GD and SARAH and uses a fixed probability $p$ (i.e., equivalent to Geom-SARAH \citep{horvath2020adaptivity}).
However, \page is more general which switches between minibatch SGD and minibatch SARAH and also allows a flexible probability $p_t$.
More importantly, the minibatch SGD update instead of GD can allow \page to solve both nonconvex finite-sum and online problems, while L2S \citep{li2020convergence} can only deal with the finite-sum case. 
Besides, our convergence analysis of \page is simple and clean, which is totally different from L2S \citep{li2020convergence}. 
Concretely, our analysis of \page directly shows the decrease for each iteration (see \eqref{eq:use-eta} or \eqref{eq:phit}), i.e., \emph{truly loopless analysis}. However, L2S \citep{li2020convergence} still uses a  \emph{double loop analysis} where they transform the probabilistic switch steps to an equivalent double loop structure and upper bound the variance term by considering all inner loop iterations together not just one iteration as ours (see Lemma 5 of L2S vs.\ our Lemma \ref{lem:var-finite}).

\section{Notation and Assumptions}
\label{sec:pre}

Let $[n]$ denote the set $\{1,2,\cdots,n\}$ and $\n{\cdot}$ denote the Euclidean norm for a vector and the spectral norm for a matrix.
Let $\inner{u}{v}$ denote the inner product of two vectors $u$ and $v$.
We use $O(\cdot)$ and $\Omega(\cdot)$ to hide the absolute constant, and $\widetilde{O}(\cdot)$ to hide the logarithmic factor. We will write $\fgap:= f(x^0) - f^*$ and $f^* := \min_{x\in \R^d} f(x)$. 

In order to prove convergence results, one usually needs the following standard assumptions depending on the setting (see e.g., \citealp{ghadimi2016mini, lei2017non, li2018simple, allen2018natasha, zhou2018stochastic, fang2018spider}).

\begin{assumption}[Bounded variance]\label{asp:bv}
	The stochastic gradient has bounded variance if $\exists \sigma >0$, such that
	\begin{equation}\label{eq:bv}
	\E_{i}[\ns{\nabla f_i(x)-\nabla f(x)}]\leq \sigma^2, \quad \forall x \in \R^d.
	\end{equation}
\end{assumption}

\begin{assumption}[Average $L$-smoothness]\label{asp:smooth}
	A function $f:\R^d\to \R$ is average $L$-smooth if $\exists L >0$, %such that
	\begin{equation}\label{eq:avgsmooth}
	\E_i[\ns{\nabla f_i(x) - \nabla f_i(y)}]\leq L^2 \ns{x-y}, ~ \forall x,y \in \R^d.
	\end{equation}
\end{assumption}

Moreover, we also prove faster linear convergence rates for nonconvex functions under the Polyak-\L{}ojasiewicz (PL) condition \citep{polyak1963gradient}.
\begin{assumption}[PL condition] \label{asp:pl}
	A function $f:\R^d\to \R$ satisfies PL condition~\footnote{It is worth noting that the PL condition does not imply convexity of $f$. For example, $f(x) = x^2 + 3\sin^2 x$ is a nonconvex function but it satisfies PL condition with $\mu=1/32$.} if $\exists \mu>0$, such that
	\begin{equation}\label{eq:pl}
	\ns{\nabla f(x)} \geq 2\mu (f(x)-f^*),~ \forall x\in \R^d.
	\end{equation}
\end{assumption}

\section{General Convergence Results}
\label{sec:result}

In this section, we present two main convergence theorems for \page (Algorithm \ref{alg:page}): i) for nonconvex finite-sum problem \eqref{prob:finite}  (Section \ref{sec:finite}), and ii) for nonconvex online problem \eqref{prob:online} (Section \ref{sec:online}).
Subsequently, we formulate several corollaries  which lead to the optimal convergence results. Finally, we provide tight lower bounds for both types of nonconvex problems to close the gap and validate the optimality of \page. 
See Table~\ref{table:nonconvex} for an overview.

\subsection{Convergence for nonconvex finite-sum problems}
\label{sec:finite}

In this section, we focus on the nonconvex finite-sum problems defined via \eqref{prob:finite}. In this case,  we do not need the bounded variance assumption (Assumption \ref{asp:bv}).

\begin{theorem}[Nonconvex finite-sum problem \eqref{prob:finite}]\label{thm:finite}
	Suppose that Assumption \ref{asp:smooth} holds. 
	Choose the stepsize 
	$\eta \leq \frac{1}{L\left(1+\sqrt{\frac{1-p}{pb'}}\right)}$,
	minibatch size $b=n$, secondary minibatch size $b'<b$, and probability $p_t\equiv p\in (0,1]$.
	Then the number of iterations performed by \page sufficient for  finding an $\epsilon$-approximate solution (i.e., $\E[\n{\nabla f(\hx_T)}]\leq \epsilon$) of nonconvex finite-sum problem \eqref{prob:finite} can be bounded by
	\begin{align*}
	T = \frac{2\fgap L}{\epsilon^2} \left( 1+ \sqrt{\frac{1-p}{pb'}}\right).
	\end{align*}
	Moreover, according to the gradient estimator of \page (Line \ref{line:grad} of Algorithm \ref{alg:page}), we know that it uses $pb + (1-p)b'$ stochastic gradients for each iteration on expectation. Thus, 
	the number of stochastic gradient computations (i.e., gradient complexity) is 
	\begin{align*}
	\#\mathrm{grad} &=b+ T\left(pb+(1-p)b'\right) \\
	&= b+ \frac{2\fgap L}{\epsilon^2} \left( 1+ \sqrt{\frac{1-p}{pb'}}\right)\left(pb+(1-p)b'\right).
	\end{align*}
	Note that the first $b$ in $\#\mathrm{grad}$ is due to the computation of $g^0$ (see Line \ref{line:sgd} in Algorithm \ref{alg:page}).
\end{theorem}

As we mentioned before, if we choose $p_t\equiv 1$ and $b=n$ (see Line \ref{line:grad} of Algorithm \ref{alg:page}),  \page reduces to the vanilla GD method. We now show that our main theorem indeed recovers the convergence result of GD.

\begin{corollary}[We recover GD by letting $p_t\equiv 1$]\label{cor:gd}
	Suppose that Assumption \ref{asp:smooth} holds. 
	Choose the stepsize 
	$\eta \leq \frac{1}{L}$,
	minibatch size $b=n$ and probability $p_t\equiv 1$.
	Then \page reduces to GD, and the number of iterations performed by \page  to find an $\epsilon$-approximate solution of the nonconvex finite-sum problem \eqref{prob:finite} can be bounded by
	$
	T = \frac{2\fgap L}{\epsilon^2}.
	$
	Moreover, the number of stochastic gradient computations (i.e., gradient complexity) is 
	\begin{align*}
	\#\mathrm{grad} = n+ \frac{2\fgap L n}{\epsilon^2} =O\left(\frac{n}{\epsilon^2}\right).
	\end{align*}
\end{corollary} 

Next, we provide a parameter setting that leads to the optimal convergence result  for nonconvex finite-sum problem \eqref{prob:finite}, which corresponds to the 6th row of Table \ref{table:nonconvex}.
Note that a fixed $p_t$ is enough for \page to obtain the optimal convergence result although people can choose different $p_t$ in practice.

\begin{corollary}[Optimal result for problem \eqref{prob:finite}]\label{cor:finite}
	Suppose that Assumption \ref{asp:smooth}  holds. 
	Choose the stepsize 
	$\eta \leq \frac{1}{L(1+\sqrt{b}/b')}$,
	minibatch size $b=n$, secondary minibatch size $b'\leq \sqrt{b}$ and probability $p_t\equiv \frac{b'}{b+b'}$.
	Then the number of iterations performed by \page  to find an $\epsilon$-approximate solution of the nonconvex finite-sum problem \eqref{prob:finite} can be bounded by
	%	\begin{align}
	$T = \frac{2\fgap L}{\epsilon^2} ( 1+ \frac{\sqrt{b}}{b'}).$
	%	\end{align}
	Moreover, the number of stochastic gradient computations (i.e., gradient complexity) is 
	\begin{align*} %\label{eq:res-finite}
	\#\mathrm{grad} %&= b+ T\left(pb+(1-p)b'\right) \\ %\leq  b+ \frac{4\fgap L}{\epsilon^2} \left( b'+ \sqrt{b}\right) 
	&\leq n + \frac{8\fgap L \sqrt{n}}{\epsilon^2}  
	= O\left(n+ \frac{\sqrt{n}}{\epsilon^2}\right).
	\end{align*}
\end{corollary} 

Finally, we establish a lower bound matching the above upper bound, which shows that the convergence result obtained by \page in Corollary \ref{cor:finite} is indeed optimal.  This lower bound corresponds to the 8th row of Table~\ref{table:nonconvex}.

\begin{theorem}[Lower bound]\label{thm:lb}
	For any $L>0$, $\fgap>0$ and $n>0$, there exists a  large enough dimension $d$ and a function $f:\R^d\to \R$ satisfying Assumption \ref{asp:smooth} in the finite-sum case  such that any linear-span first-order algorithm needs $\Omega(n + \frac{\fgap L \sqrt{n}}{\epsilon^2})$ stochastic gradient computations in order to finding an $\epsilon$-approximate solution, i.e., a point $\hx$ such that $\E\n{\nabla f(\hx)} \leq \epsilon$.
\end{theorem}

\subsection{Convergence for nonconvex online problems}
\label{sec:online}
In this section, we focus on the nonconvex online problems, i.e., \eqref{prob:online}.
Recall that we refer this online problem \eqref{prob:online} as the finite-sum problem \eqref{prob:finite} with large or infinite $n$. Also, we need the bounded variance assumption (Assumption \ref{asp:bv}) in this online case.
Similarly, we first present the main theorem in this online case and then provide corollaries with the optimal convergence results. Finally, we provide tight lower bound for validating the optimality of \page.

\begin{theorem}[Nonconvex online problem \eqref{prob:online}]\label{thm:online}
	Suppose that Assumptions \ref{asp:bv} and \ref{asp:smooth} hold. 
	Choose the stepsize 
	$\eta \leq \frac{1}{L\left(1+\sqrt{\frac{1-p}{pb'}}\right)}$,
	minibatch size $b=\min \{\lceil \frac{2\sigma^2}{\epsilon^2} \rceil, n\}$, secondary minibatch size $b'<b$ and probability $p_t\equiv p\in (0,1]$.
	Then the number of iterations performed by \page  to find an $\epsilon$-approximate solution ($\E[\n{\nabla f(\hat{x}_T)}]\leq \epsilon$) of nonconvex online problem \eqref{prob:online} can be bounded by
	\begin{align*}
	T = \frac{4\fgap L}{\epsilon^2} \left( 1+ \sqrt{\frac{1-p}{pb'}}\right) + \frac{1}{p}.
	\end{align*}
	Moreover, %according to the gradient estimator of \page (Line \ref{line:grad} of Algorithm \ref{alg:page}), we know that it uses $pb + (1-p)b'$ stochastic gradients for each iteration on the expectation. Thus, 
	the number of stochastic gradient computations (gradient complexity) $\#\mathrm{grad} =b+ T\left(pb+(1-p)b'\right)$  is 
	\begin{align*}
	2b+ \frac{(1-p)b'}{p} + \frac{4\fgap L}{\epsilon^2} \left( 1+ \sqrt{\frac{1-p}{pb'}}\right)\left(pb+(1-p)b'\right).
	\end{align*}
\end{theorem}

Similarly, if we choose $p_t\equiv 1$ (see Line \ref{line:grad} of Algorithm \ref{alg:page}), the \page method reduces to the vanilla minibatch SGD method.
Here we theoretically show that our main theorem with $p_t\equiv 1$ can recover the convergence result of SGD in the following Corollary \ref{cor:sgd}.

\begin{corollary}[We recover SGD by letting $p_t\equiv 1$]\label{cor:sgd}
	Suppose that Assumptions \ref{asp:bv} and \ref{asp:smooth} hold. 
	Let stepsize 
	$\eta \leq \frac{1}{L}$,
	minibatch size $b=\lceil \frac{2\sigma^2}{\epsilon^2} \rceil$ and probability $p_t\equiv 1$,
	then the number of iterations performed by \page to find an $\epsilon$-approximate solution of nonconvex online problem \eqref{prob:online}  can be bounded by
	$
	T = \frac{4\fgap L}{\epsilon^2} +1.
	$ 
	Moreover, the number of stochastic gradient computations (gradient complexity) is 
	\begin{align*}
	\#\mathrm{grad} = \frac{4\sigma^2}{\epsilon^2}+ \frac{8\fgap L \sigma^2}{\epsilon^4}  = O\left(\frac{\sigma^2}{\epsilon^4}\right).
	\end{align*}
\end{corollary}

Now, we provide a parameter setting that leads to the optimal convergence result of our main theorem for nonconvex online problem \eqref{prob:online}, which corresponds to the 14th row of Table \ref{table:nonconvex}.
Similarly, a fixed $p_t$ is enough for \page to obtain the optimal convergence result in this online case.
\begin{corollary}[Optimal result for problem \eqref{prob:online}]\label{cor:online}
	Suppose that Assumptions \ref{asp:bv} and \ref{asp:smooth} hold. 
	Choose the stepsize 
	$\eta \leq \frac{1}{L(1+\sqrt{b}/b')}$,
	minibatch size $b=\min \{\lceil \frac{2\sigma^2}{\epsilon^2} \rceil, n\}$, secondary minibatch size $b'\leq \sqrt{b}$ and probability $p_t\equiv \frac{b'}{b+b'}$.
	Then the number of iterations performed by \page sufficient to find an $\epsilon$-approximate solution of nonconvex online problem \eqref{prob:online} can be bounded by
	%	\begin{align}
	$T = \frac{4\fgap L}{\epsilon^2} ( 1+ \frac{\sqrt{b}}{b'}) +\frac{b+b'}{b'}.$
	%	\end{align}
	Moreover, 
	%	according to the gradient estimator of \page (Line \ref{line:grad} of Algorithm \ref{alg:page}), we know that it uses $pb + (1-p)b'=\frac{2bb'}{b+b'}<2b'$ stochastic gradients for each iteration on the expectation. Thus, 
	the number of stochastic gradient computations (i.e., gradient complexity) is 
	\begin{align*}
	\#\mathrm{grad} %&= b+ T\left(pb+(1-p)b'\right) \\%\leq  b+ \frac{8\fgap L}{\epsilon^2} \left( b'+ \sqrt{b}\right) 
	&\leq 3b+ \frac{16\fgap L \sqrt{b}}{\epsilon^2}
	=O\Big(b + \frac{\sqrt{b}}{\epsilon^2}\Big).
	\end{align*}
\end{corollary}

Before we provide our lower bound, we first recall the lower bound established by \citet{arjevani2019lower}.

\begin{theorem}[\citealp{arjevani2019lower}]\label{thm:lbonline}
	For any $L>0$, $\fgap>0$ and $\sigma^2>0$, there exists a large enough dimension $d$ and function $f:\R^d\to \R$ satisfying Assumptions \ref{asp:bv} and \ref{asp:smooth} in the online case (here $n$ is infinite) such that any linear-span first-order algorithm needs $\Omega(\frac{\sigma^2}{\epsilon^2} + \frac{\fgap L \sigma}{\epsilon^3})$ stochastic gradient computations in order to find an $\epsilon$-approximate solution, i.e., a point $\hx$ such that $\E\n{\nabla f(\hx)} \leq \epsilon$.
\end{theorem}

Now, we provide a lower bound corollary which directly follows from the lower bound Theorem \ref{thm:lbonline} given by \citet{arjevani2019lower} and our Theorem \ref{thm:lb}.
It  indicates that the convergence result obtained by \page in Corollary \ref{cor:online} is indeed optimal. % and corresponds to the last row of Table \ref{table:nonconvex}.
\begin{corollary}[Lower bound]\label{cor:lbonline}
	For any $L>0$, $\fgap>0$, $\sigma^2>0$ and $n>0$, there exists a large enough dimension $d$ and a function $f:\R^d\to \R$ satisfying Assumptions \ref{asp:bv} and \ref{asp:smooth} in the online case (here $n$ may be finite) such that any linear-span first-order algorithm needs $\Omega(b + \frac{\fgap L \sqrt{b}}{\epsilon^2})$, where $b = \min\{\frac{\sigma^2}{\epsilon^2}, n\}$,  stochastic gradient computations for finding an $\epsilon$-approximate solution, i.e., a point $\hx$ such that $\E\n{\nabla f(\hx)} \leq \epsilon$.
\end{corollary}

\section{Better Convergence under PL Condition}
\label{sec:pl}
In this section, we show that better convergence can be achieved if the loss function $f$ satisfies the PL condition (Assumption~\ref{asp:pl}).
Note that under the PL condition, one can obtain a faster linear convergence $O(\cdot\log \frac{1}{\epsilon})$ (see Corollary~\ref{cor:finite-pl}) rather than the sublinear convergence $O(\cdot\frac{1}{\epsilon^2})$ (see Corollary~\ref{cor:finite}).
In many cases, although the loss function $f$ is globally nonconvex, some local regions (e.g., large gradient regions) may satisfy the PL condition. We prove that \page can \emph{automatically} switch to the faster convergence rate in these regions where $f$ satisfies PL condition locally. 

As in Section \ref{sec:result}, here we also establish two main theorems and the deduce corollaries for both finite-sum and online regimes. The convergence results are also listed in Table \ref{table:nonconvex} (i.e., the middle row and last row).

\begin{theorem}[Nonconvex finite-sum problem~\eqref{prob:finite} under PL]\label{thm:finite-pl} 
	Suppose that Assumptions \ref{asp:smooth} and \ref{asp:pl} hold. 
	Choose the stepsize 
	$\eta \leq \min\{ \frac{1}{L\left(1+\sqrt{\frac{1-p}{pb'}}\right)},~  \frac{p}{2\mu}  \}$,
	minibatch size $b=n$, secondary minibatch size $b'<b$, and probability $p_t\equiv p\in (0,1]$.
	Then the number of iterations performed by \page sufficient for  finding an $\epsilon$-solution ($\E[f(x^T)-f^*]\leq \epsilon$) of nonconvex finite-sum problem \eqref{prob:finite} can be bounded by
	\begin{align*}
	\compactify	T =  \left(\left(1+\sqrt{\frac{1-p}{pb'}}\right) \kappa + \frac{2}{p} \right) \log \frac{\fgap}{\epsilon}.
	\end{align*}
	Moreover, %according to the gradient estimator of \page (Line \ref{line:grad} of Algorithm \ref{alg:page}), we know that it uses $pb + (1-p)b'$ stochastic gradients for each iteration on the expectation. Thus, 
	the number of stochastic gradient computations (i.e., gradient complexity) $\#\mathrm{grad} =b+ T\left(pb+(1-p)b'\right)$  is 
	\begin{align*}
	b+  \left(pb+(1-p)b'\right) \left(\left(1+\sqrt{\frac{1-p}{pb'}}\right) \kappa + \frac{2}{p} \right)\log \frac{\fgap}{\epsilon}.
	\end{align*}
\end{theorem}

\begin{corollary}\label{cor:finite-pl}
	Suppose that Assumptions \ref{asp:smooth} and \ref{asp:pl} hold. 
	Let stepsize $\eta \leq \min \{ \frac{1}{L(1+\sqrt{b}/b')},~  \frac{b'}{2\mu (b+b')}  \}$,
	minibatch size $b=n$, secondary minibatch size $b'\leq \sqrt{b}$, and probability $p_t\equiv \frac{b'}{b+b'}$.
	Then the number of iterations performed by \page to find an $\epsilon$-solution of nonconvex finite-sum problem \eqref{prob:finite} can be bounded by
	%	\begin{align}
	$T = \left((1+\frac{\sqrt{b}}{b'}) \kappa + \frac{2(b+b')}{b'} \right)  \log \frac{\fgap}{\epsilon}.$
	%	\end{align}
	Moreover, the number of stochastic gradient computations (gradient complexity) is 
	\begin{align*} 
	\#\mathrm{grad} %&= b+ T\left(pb+(1-p)b'\right) \\
%	&\leq n+ (4\sqrt{n}\kappa + 4n) \log \frac{\fgap}{\epsilon} \\
	&= O\left((n + \sqrt{n}\kappa)\log \frac{1}{\epsilon} \right).
	\end{align*}
\end{corollary} 

{\noindent\bf Remark:} Note that Corollary \ref{cor:finite-pl} uses exactly the same parameter setting as in Corollary \ref{cor:finite} in the large condition number case (i.e., $\kappa:=\frac{L}{\mu} \geq 2\sqrt{n}$, then the stepsize turns to $\eta \leq \frac{1}{L(1+\sqrt{b}/b')}$).
Thus, \page  can automatically switch to this faster linear convergence rate $O(\cdot\log\frac{1}{\epsilon})$ instead of the sublinear convergence $O(\frac{\cdot}{\epsilon^2})$ in Corollary \ref{cor:finite} in some regions where $f$ satisfies the PL condition locally.

\begin{figure*}[!t]
	\centering
	\begin{minipage}{0.34\textwidth}
		\centering
		\includegraphics[width=\linewidth]{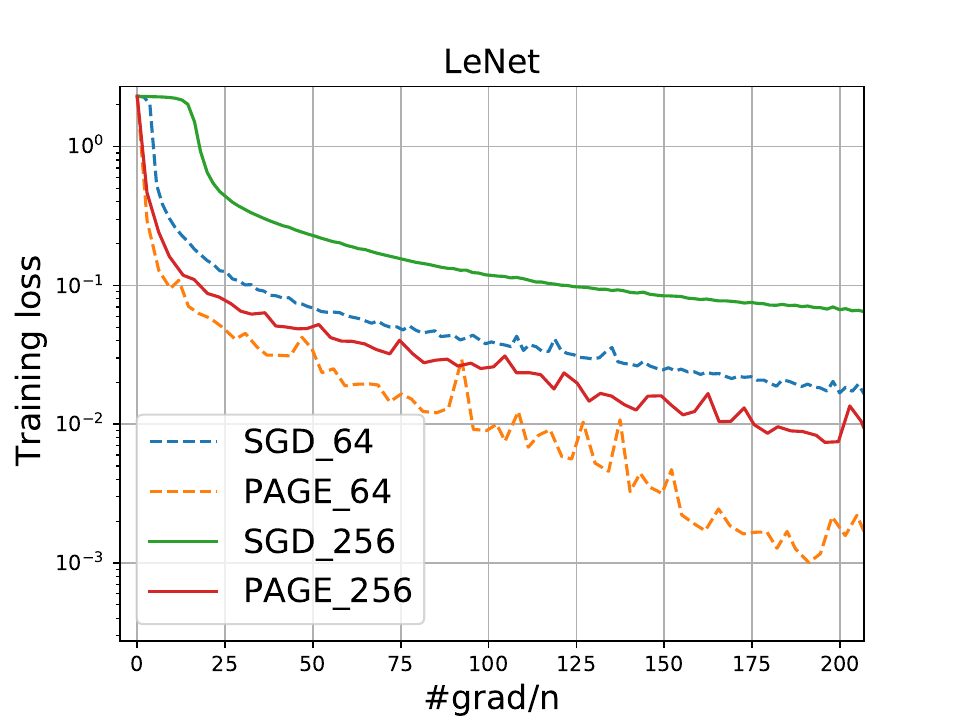}%\hspace{1mm}
	\end{minipage}%\hspace{mm}
	\begin{minipage}{0.34\textwidth}
		\centering
		\includegraphics[width=\linewidth]{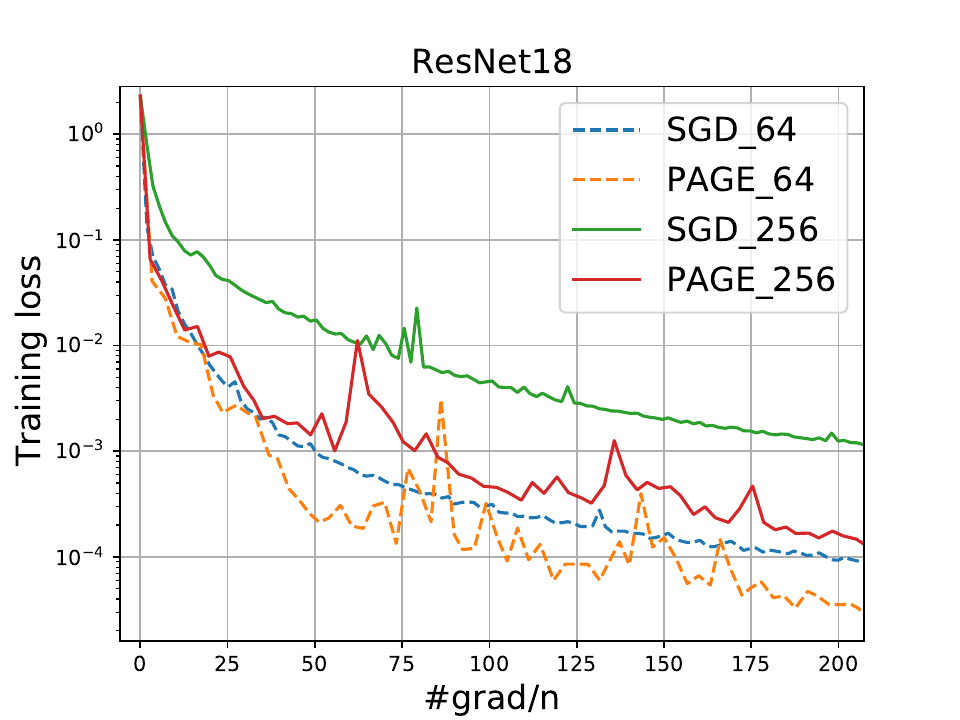}%\hspace{1mm}
	\end{minipage}%\hspace{mm}
	\begin{minipage}{0.34\textwidth}
		\centering
		\includegraphics[width=\linewidth]{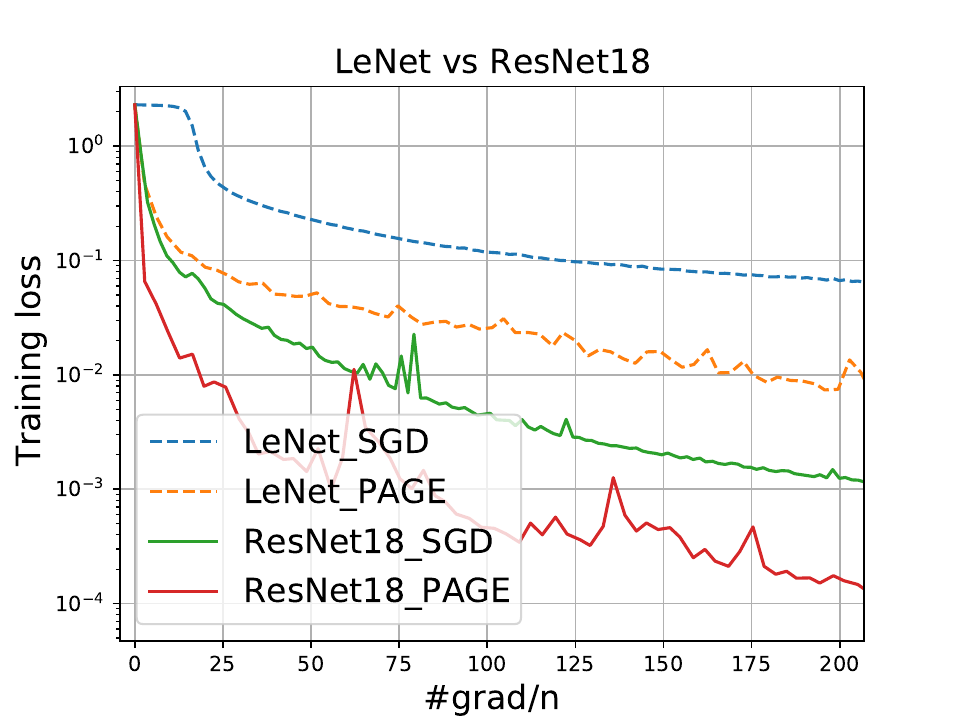}%\hspace{1mm}
	\end{minipage}\\%\hspace{mm}
	\begin{minipage}{0.34\textwidth}
		\centering
		\includegraphics[width=\linewidth]{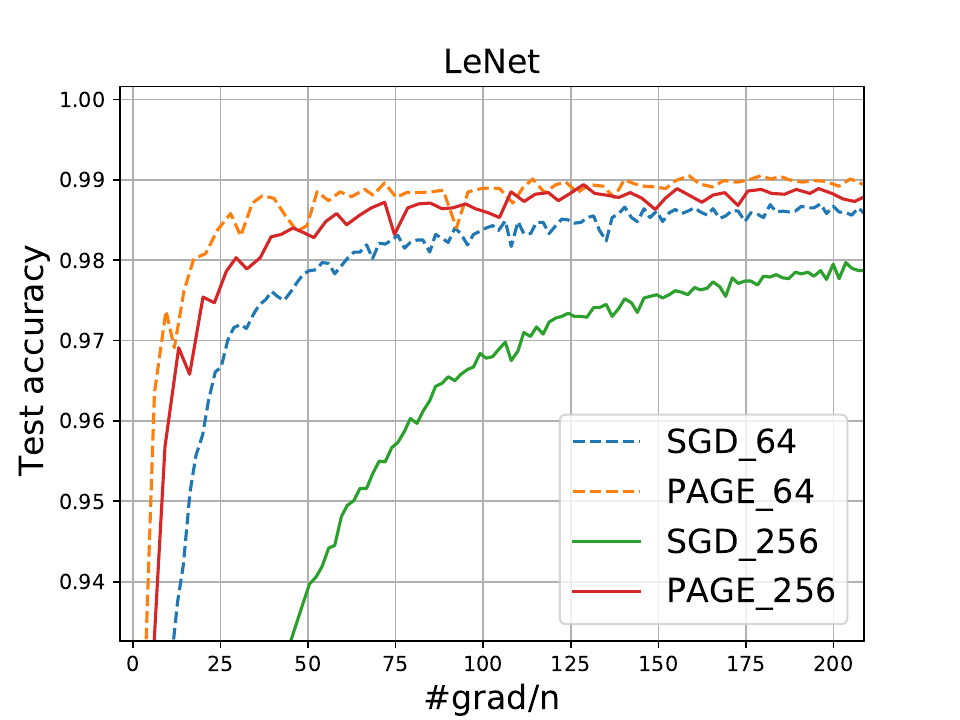}%\hspace{1mm}
		%		\caption{MNIST}
		%		\label{fig:letnet-mnist}
		
		(1a) Different minibatch size $b$ 
	\end{minipage}%\hspace{mm}
	\begin{minipage}{0.34\textwidth}
		\centering
		\includegraphics[width=\linewidth]{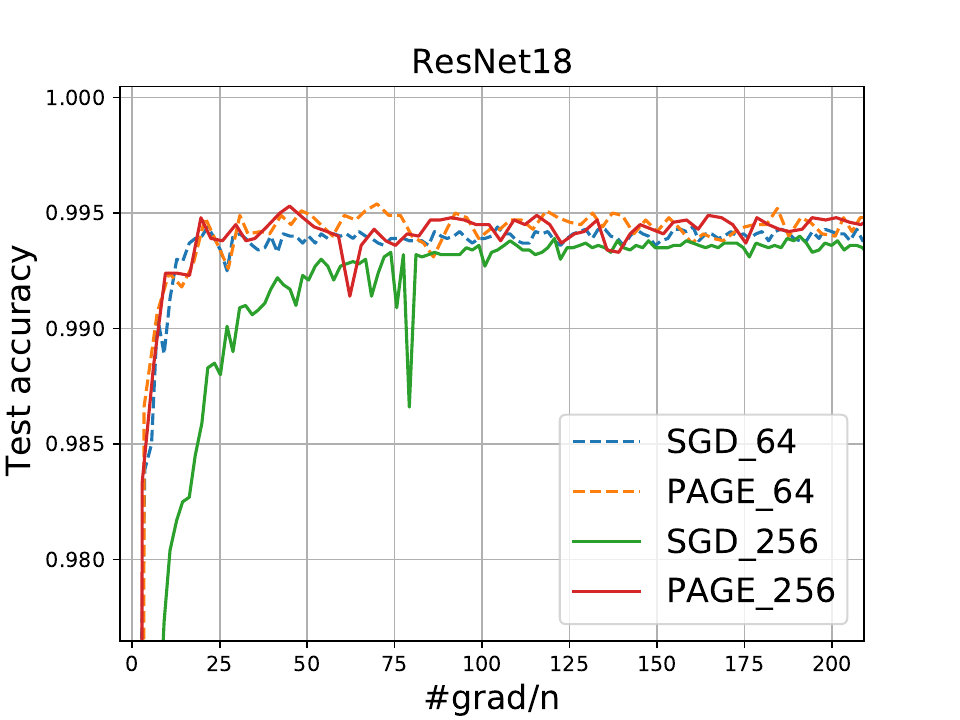}%\hspace{1mm}
		%		\caption{MNIST}
		%		\label{fig:resnet-mnist}
		
		(1b) Different minibatch size $b$ 
	\end{minipage}%\hspace{mm}
	\begin{minipage}{0.34\textwidth}
		\centering
		\includegraphics[width=\linewidth]{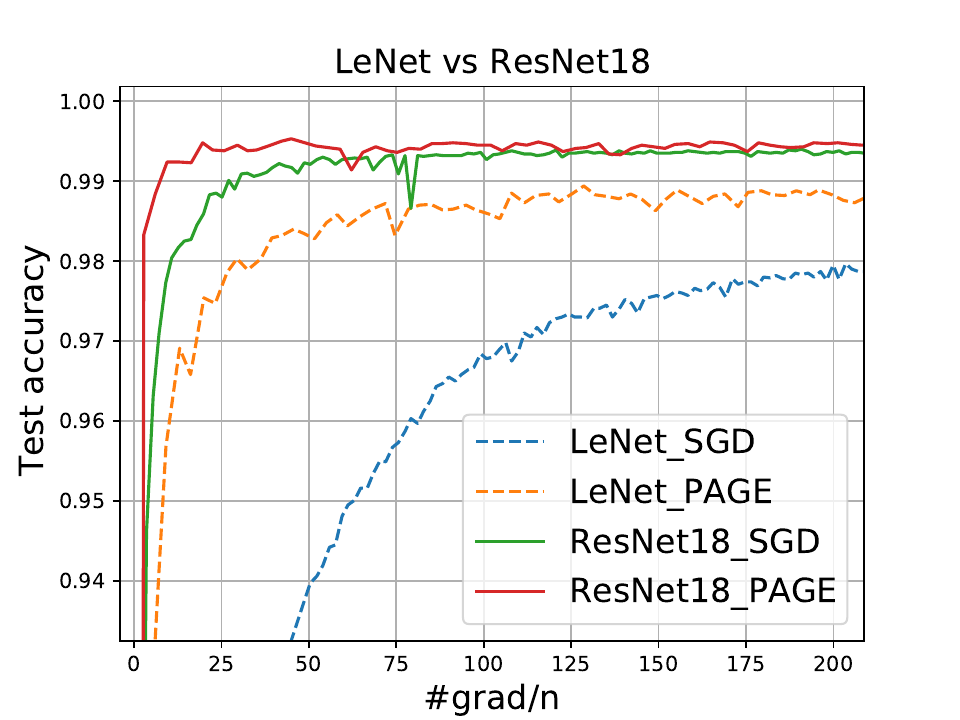}%\hspace{1mm}
		%		\caption{MNIST}
		%		\label{fig:lenet-resnet-mnist}
		
		(1c) Different neural networks
	\end{minipage}
	\vspace{-2mm}
	\caption{LeNet and ResNet18 on MNIST dataset}
	\label{fig:mnist}
\end{figure*}

\begin{theorem}[Nonconvex online problem~\eqref{prob:online} under PL]\label{thm:online-pl}
	Suppose that Assumptions \ref{asp:bv}, \ref{asp:smooth} and \ref{asp:pl} hold. 
	Choose the stepsize 
	$\eta \leq \min\{ \frac{1}{L\left(1+\sqrt{\frac{1-p}{pb'}}\right)},~  \frac{p}{2\mu}  \}$,
	minibatch size $b=\min \{\lceil \frac{2\sigma^2}{\mu \epsilon} \rceil, n\}$, secondary minibatch size $b'<b$, and probability $p_t\equiv p\in (0,1]$.
	Then the number of iterations performed by \page sufficient for  finding an $\epsilon$-solution ($\E[f(x^T)-f^*]\leq \epsilon$) of nonconvex finite-sum problem \eqref{prob:finite} can be bounded by
	\begin{align*}
	\compactify	T =  \left(\Big(1+\sqrt{\frac{1-p}{pb'}}\Big) \kappa + \frac{2}{p} \right) \log \frac{2 \fgap}{\epsilon}.
	\end{align*}
	Moreover, %according to the gradient estimator of \page (Line \ref{line:grad} of Algorithm \ref{alg:page}), we know that it uses $pb + (1-p)b'$ stochastic gradients for each iteration on the expectation. Thus, 
	the number of stochastic gradient computations (i.e., gradient complexity) $\#\mathrm{grad} =b+ T\left(pb+(1-p)b'\right)$ is 
	\begin{align*}
	b+  \left(pb+(1-p)b'\right)\left(\Big(1+\sqrt{\frac{1-p}{pb'}}\Big) \kappa + \frac{2}{p} \right) \log \frac{2\fgap}{\epsilon}.
	\end{align*}
\end{theorem}

\begin{corollary}\label{cor:online-pl}
	Suppose that Assumptions \ref{asp:bv}, \ref{asp:smooth} and \ref{asp:pl} hold. 
	Choose the stepsize 
	$\eta \leq \min \{ \frac{1}{L(1+\sqrt{b}/b')},~  \frac{b'}{2\mu (b+b')}  \}$,
	minibatch size $b=\min \{\lceil \frac{2\sigma^2}{\mu \epsilon} \rceil, n\}$, secondary minibatch $b'\leq \sqrt{b}$ and probability $p_t\equiv \frac{b'}{b+b'}$.
	Then the number of iterations performed by \page  to find an $\epsilon$-solution of nonconvex online problem  \eqref{prob:online} can be bounded by
	%	\begin{align}
	$T =  \left((1+\frac{\sqrt{b}}{b'}) \kappa + \frac{2(b+b')}{b'} \right)  \log \frac{2\fgap}{\epsilon}.$
	%	\end{align}
	Moreover, the number of stochastic gradient computations (gradient complexity) is
	\vspace{-2mm}
	\begin{align*}
	\#\mathrm{grad} %= b+ T\left(pb+(1-p)b'\right) %\\
	%&\leq b+ (4\sqrt{b}\kappa + 4b)\log \tfrac{2 \fgap}{\epsilon}
	= O\left((b + \sqrt{b}\kappa)\log \frac{1}{\epsilon} \right).
	\end{align*}
\end{corollary}

\section{Experiments}

In this section, we conduct several deep learning experiments for multi-class image classification. Concretely, we compare our \page algorithm with vanilla SGD by running standard LeNet \citep{lecun1998gradient}, VGG \citep{simonyan2014very} and ResNet \citep{he2016deep} models on MNIST  \citep{lecun1998gradient} and CIFAR-10 \citep{krizhevsky2009learning} datasets.
We implement the algorithms in PyTorch \citep{paszke2019pytorch} and run the experiments on several NVIDIA Tesla V100 GPUs.

According to the update form in \page (see Line \ref{line:grad} of Algorithm \ref{alg:page}),  \page enjoys a lower computational cost than vanilla minibatch SGD (i.e., $p_t\equiv 1$ in \page) since $b'<b$.  Thus, in the experiments we want to show how the performance of \page compares with vanilla minibatch SGD under different minibatch sizes $b$ (i.e., $b=64, 256, 512$). Note that we do not tune the parameters for \page, i.e., we  set $b'=\sqrt{b}$ and $p_t\equiv \frac{b'}{b+b'}=\frac{\sqrt{b}}{b+\sqrt{b}}$ according to our theoretical results (see e.g., Corollary~\ref{cor:finite} and \ref{cor:online}). For the stepsize/learning rate $\eta$, we choose the same one for both \page and minibatch SGD according to the theoretical results.

\begin{figure*}[t]
	\centering
	\begin{minipage}{0.34\textwidth}
		\centering
		\includegraphics[width=\linewidth]{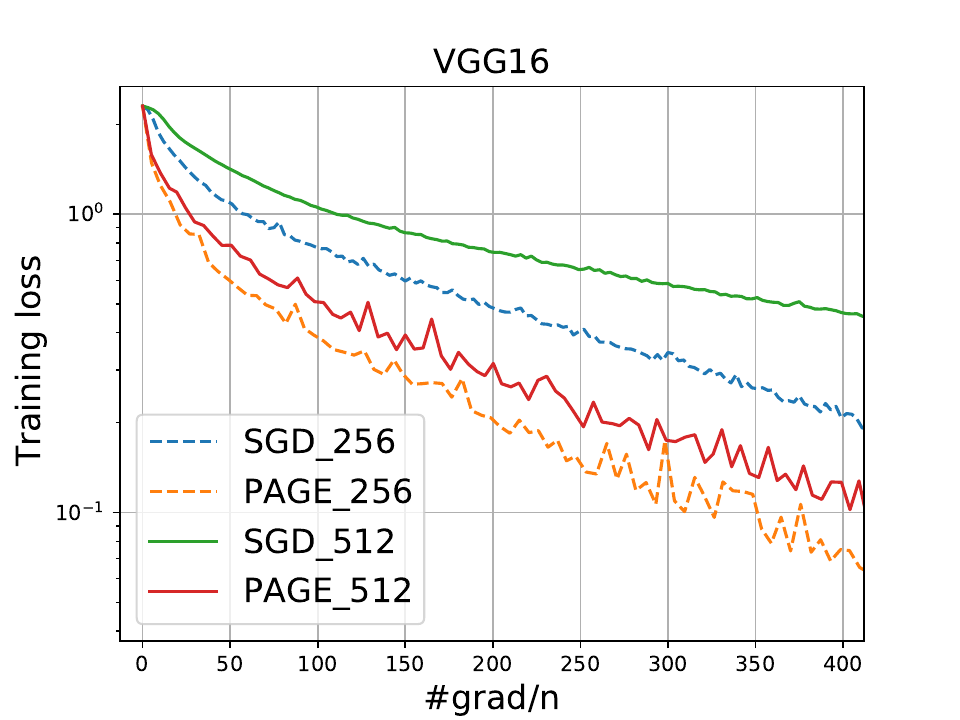}%\hspace{1mm}
	\end{minipage}%\hspace{mm}
	\begin{minipage}{0.34\textwidth}
		\centering
		\includegraphics[width=\linewidth]{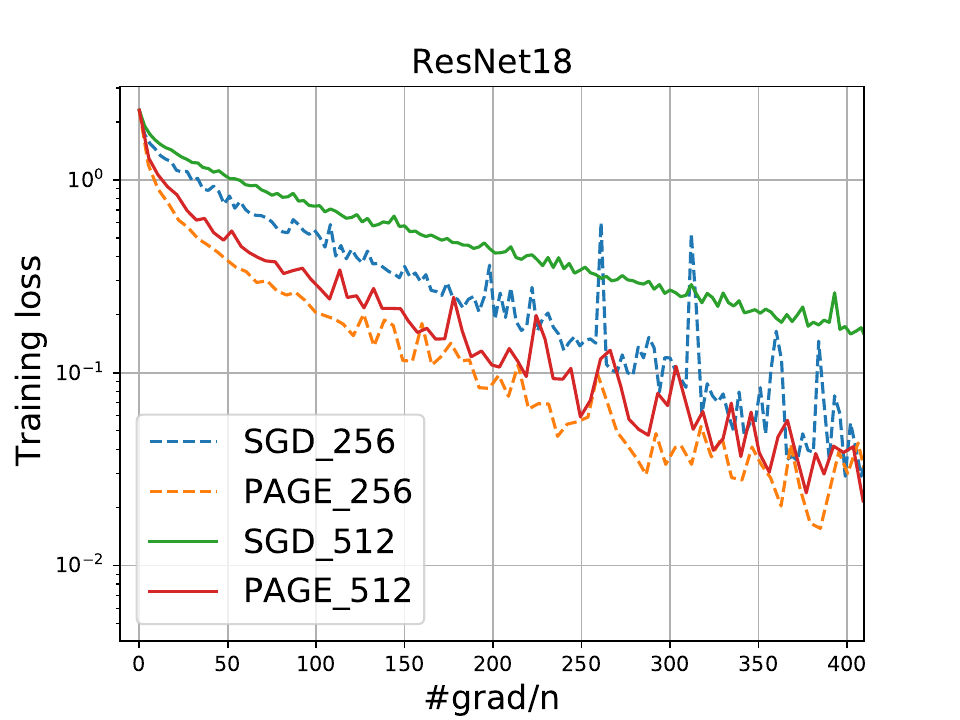}%\hspace{1mm}
	\end{minipage}%\hspace{mm}
	\begin{minipage}{0.34\textwidth}
		\centering
		\includegraphics[width=\linewidth]{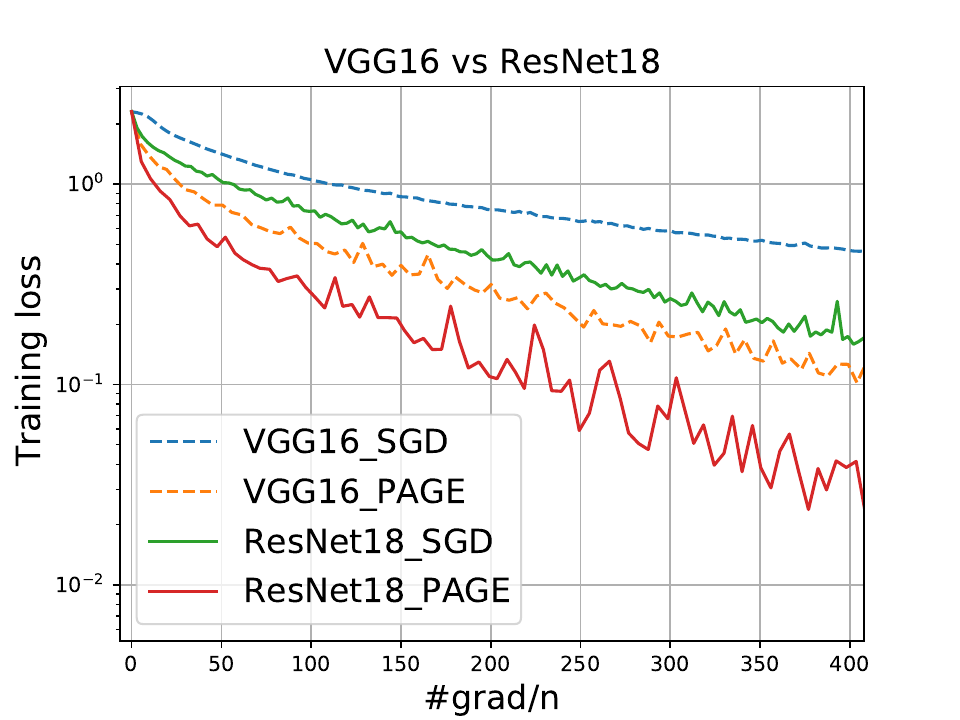}%\hspace{1mm}
	\end{minipage}\\%\hspace{mm}
	\begin{minipage}{0.34\textwidth}
		\centering
		\includegraphics[width=\linewidth]{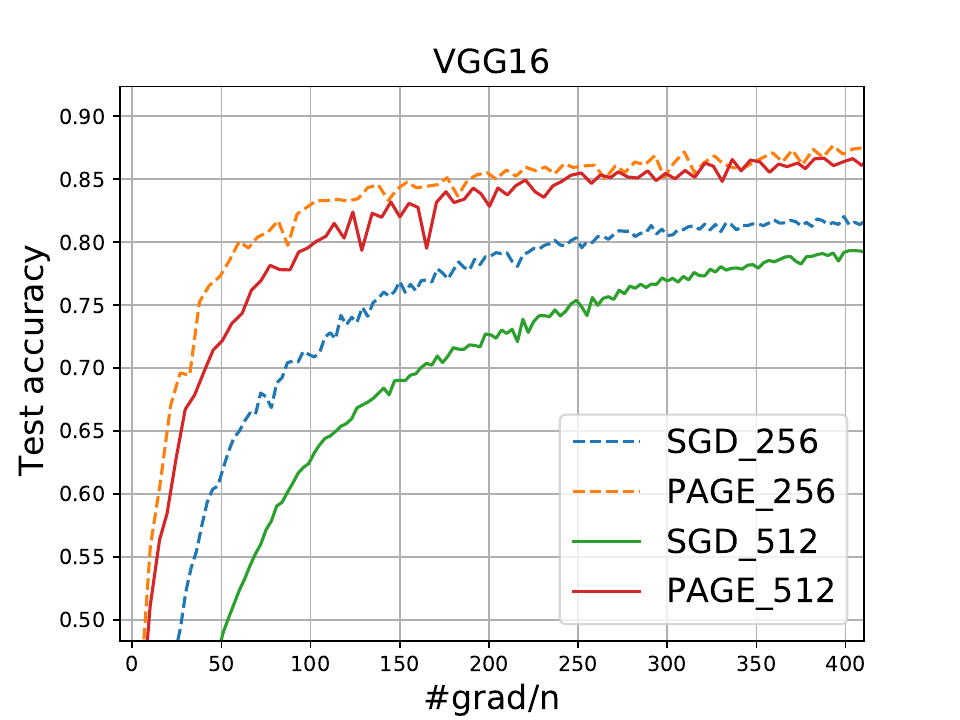}%\hspace{1mm}
		%		\caption{VGG16 on CIFAR-10}
		%		\label{fig:letnet-cifar10}
		
		(2a) Different minibatch size $b$
	\end{minipage}%\hspace{mm}
	\begin{minipage}{0.34\textwidth}
		\centering
		\includegraphics[width=\linewidth]{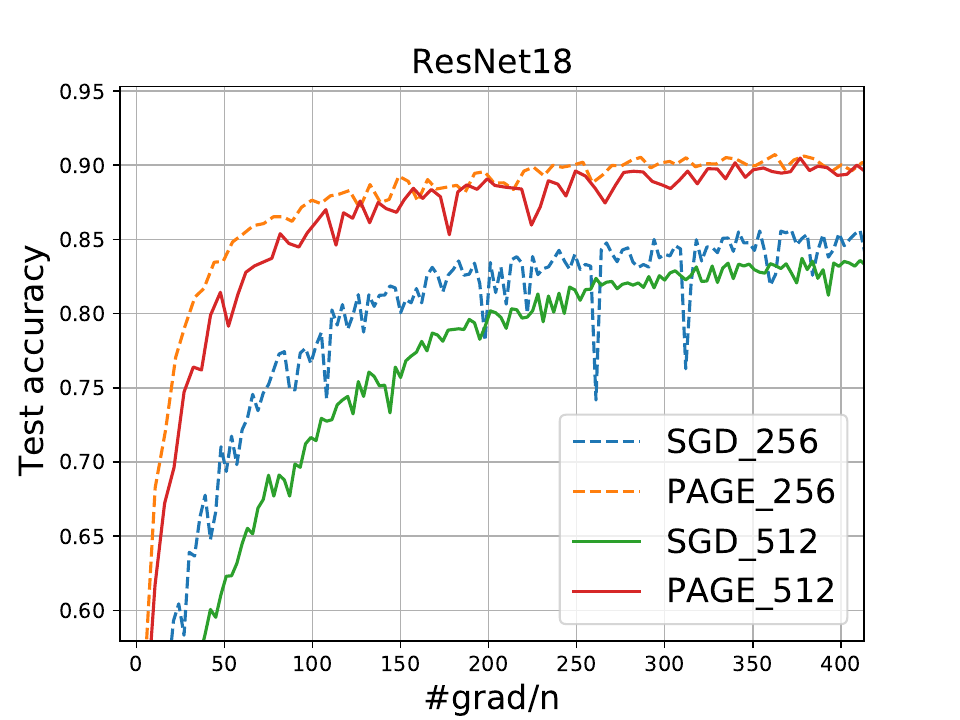}%\hspace{1mm}
		%		\caption{ResNet18 on CIFAR-10}
		%		\label{fig:resnet-cifar10}
		
		(2b) Different minibatch size $b$
	\end{minipage}%\hspace{mm}
	\begin{minipage}{0.34\textwidth}
		\centering
		\includegraphics[width=\linewidth]{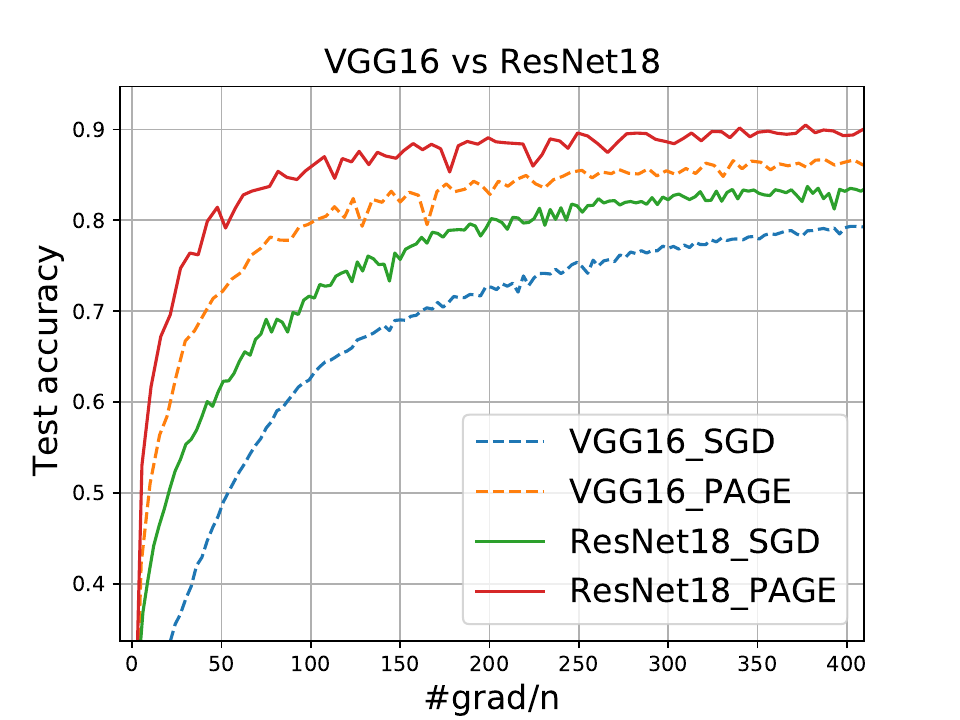}%\hspace{1mm}
		%		\caption{VGG16 vs ResNet18 on CIFAR-10}
		%		\label{fig:lenet-resnet-cifar10}
		
		(2c) Different neural networks
	\end{minipage}
	\vspace{-2mm}
	\caption{VGG16 and ResNet18 on CIFAR-10 dataset}
	\label{fig:cifar10}
\end{figure*}

Concretely, in Figure \ref{fig:mnist}, we choose standard minibatch $b=64$ and $b=256$ for both \page and vanilla minibatch SGD for MNIST experiments. In Figure \ref{fig:cifar10}, we choose $b=256$ and $b=512$ for CIFAR-10 experiments.
The first row of Figures \ref{fig:mnist} and \ref{fig:cifar10} denotes the training loss with respect to the gradient computations, and the second row denotes the test accuracy with respect to the gradient computations.
Both Figures \ref{fig:mnist} and \ref{fig:cifar10} demonstrate that \page not only converges much faster than SGD in training but also achieves higher test accuracy (which is typically very important in practice, e.g., lead to a better model). 
Moreover, the performance gap between \page and SGD is larger when the minibatch size $b$ is larger (i.e, gap between solid lines in Figures 1a, 1b, 2a, 2b), which is consistent with the update form of \page, i.e, it reuses the previous gradient with a small adjustment (lower computational cost $b'=\sqrt{b}$ instead of $b$) with probability $1-p_t$.
The experimental results validate our theoretical results and confirm the practical superiority of \page.

\begin{figure}[!t]
	\centering
	\includegraphics[width=0.51\linewidth]{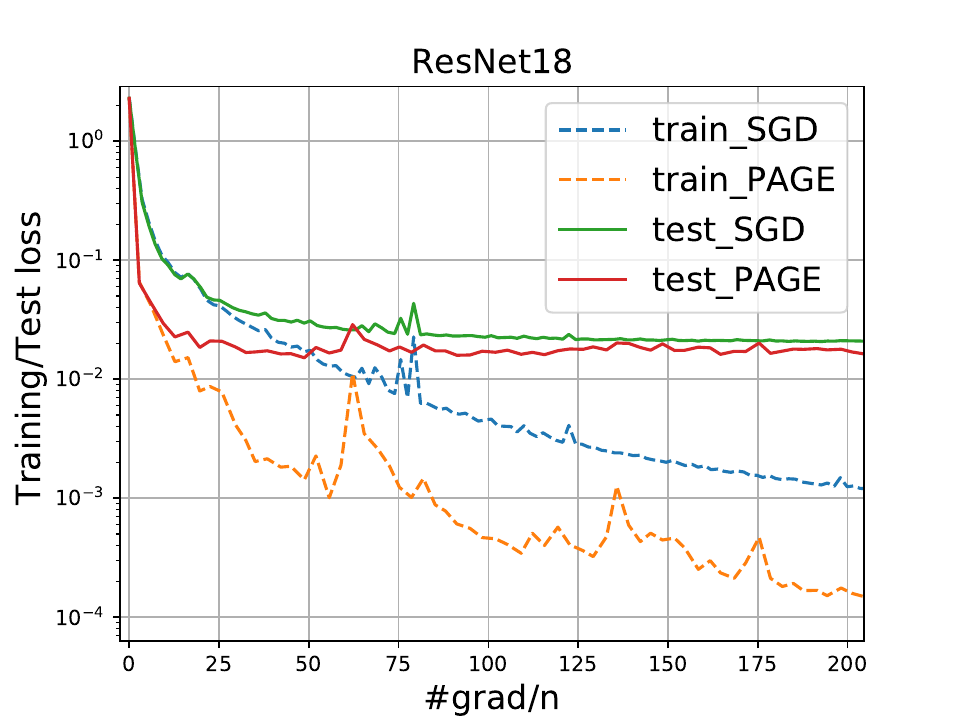}%\hspace{1mm}
	\includegraphics[width=0.51\linewidth]{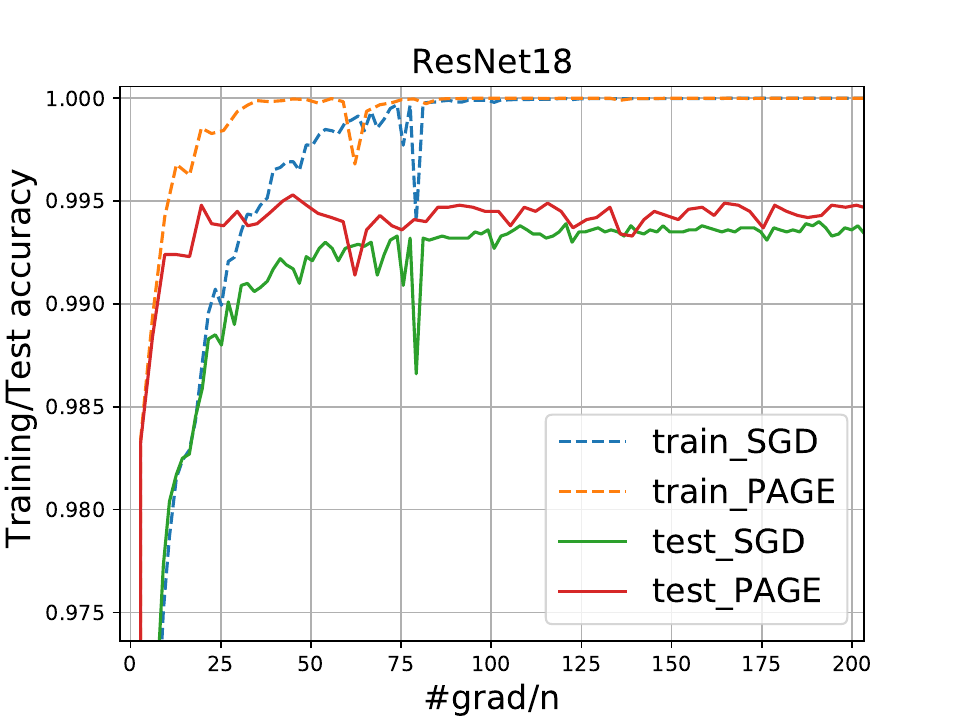}%\hspace{1mm}
	
	~(3a) Training/test loss ~~
	(3b) Training/test accuracy
	
		\vspace{-2mm}
	\caption{ResNet18 on MNIST dataset}
	\label{fig:mnist-train-test}
\end{figure}

\begin{figure}[!t]
%	\vspace{5mm}
	\centering
	\includegraphics[width=0.51\linewidth]{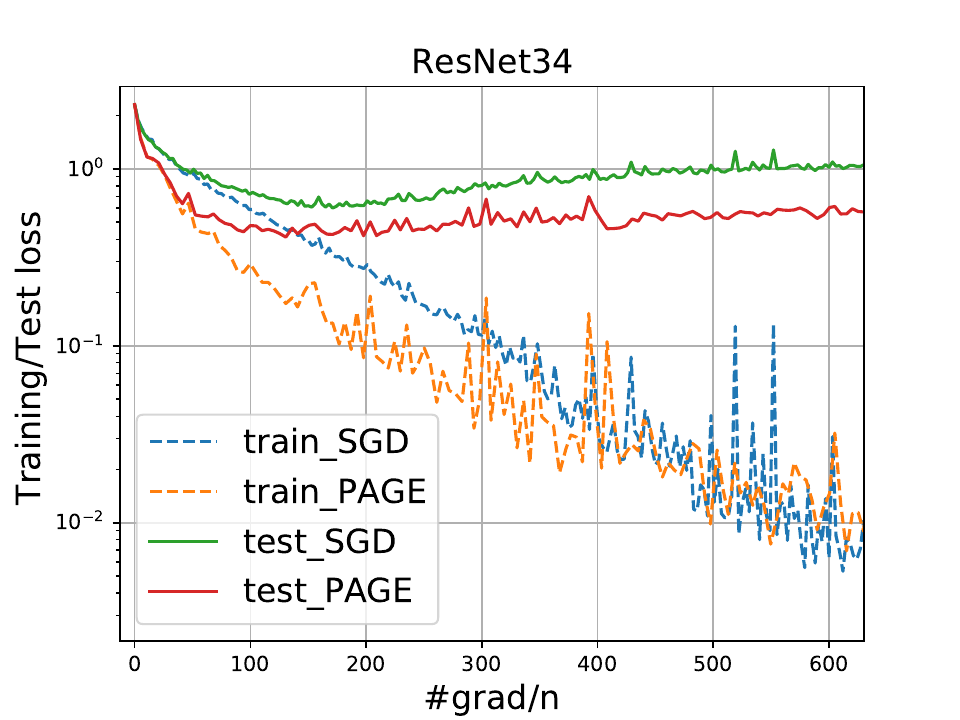}%\hspace{1mm}
	\includegraphics[width=0.51\linewidth]{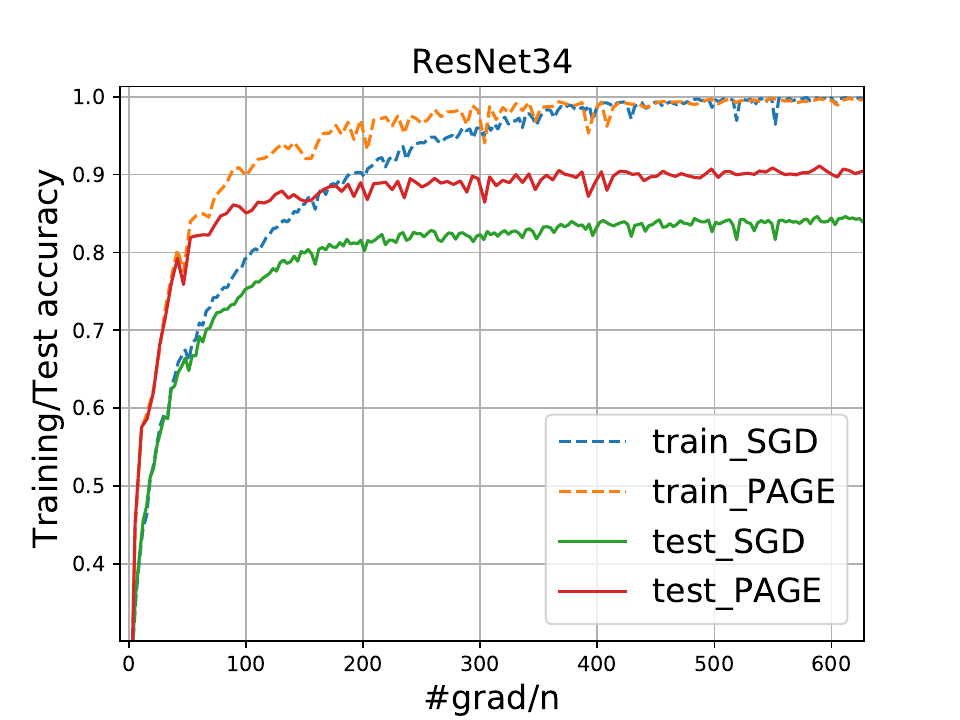}%\hspace{1mm}
	
	~(4a) Training/test loss ~~
	(4b) Training/test accuracy
	
		\vspace{-2mm}
	\caption{ResNet34 on CIFAR-10 dataset}
	\label{fig:cifar10-train-test}
		\vspace{-1mm}
\end{figure}

In the following, we conduct extra experiments for comparing the training loss and test loss (Figure 3a, 4a), and training accuracy and test accuracy (Figure 3b, 4b) between \page and SGD.
Note that Figure \ref{fig:mnist-train-test} (i.e., 3a, 3b) uses MNIST dataset and Figure \ref{fig:cifar10-train-test} (i.e., 4a, 4b) uses CIFAR-10 dataset.
Figures (3a) and (4a) also demonstrate that \page converges much faster than SGD both in training loss and test loss.
Moreover, Figures (3b) and (4b) demonstrate that \page achieves the higher test accuracy than SGD and converges faster in training accuracy.
Thus, our \page is not only converging faster than SGD in training but also achieves the higher test accuracy (which is typically very important in practice, e.g., lead to a better model).
Again, the experimental results validate our theoretical results and confirm the practical superiority of \page.

\section{Conclusion}
In this paper, we propose a simple and optimal \page algorithm for both nonconvex finite-sum and online optimization.
We prove tight lower bounds and show that \page achieves the optimal convergence results matching our lower bounds for both nonconvex finite-sum problems and online problems.
We also show that for nonconvex functions satisfying the PL condition, \page can automatically switch to a faster linear convergence rate.
Besides, \page is easy to implement and we conduct several deep learning experiments (e.g., LeNet, VGG, ResNet) in PyTorch which confirm the practical superiority of \page.
More importantly, the novel convergence analysis of \page is very simple and clean. Thus \page and its analysis can be easily adopted and generalized to other works. In fact, it already leads to some further breakthroughs in communication-efficient distributed learning (e.g., \citealp{gorbunov2021marina, peter2021ef}).

% Acknowledgements should only appear in the accepted version.
\section*{Acknowledgements}
Zhize Li and Peter Richt\'{a}rik were supported by KAUST Baseline Research Fund. 
Hongyan Bao and Xiangliang Zhang were supported by KAUST Competitive Research Grant URF/1/3756-01-01.

% In the unusual situation where you want a paper to appear in the
% references without citing it in the main text, use \nocite
%\nocite{langley00}
\bibliography{page}
\bibliographystyle{icml2021}

%%%%%%%%%%%%%%%%%%%%%%%%%%%%%%%%%%%%%%%%%%%%%%%%%%%%%%%%%%%%%%%%%%%%%%%%%%%%%%%
%%%%%%%%%%%%%%%%%%%%%%%%%%%%%%%%%%%%%%%%%%%%%%%%%%%%%%%%%%%%%%%%%%%%%%%%%%%%%%%
% DELETE THIS PART. DO NOT PLACE CONTENT AFTER THE REFERENCES!
%%%%%%%%%%%%%%%%%%%%%%%%%%%%%%%%%%%%%%%%%%%%%%%%%%%%%%%%%%%%%%%%%%%%%%%%%%%%%%%
%%%%%%%%%%%%%%%%%%%%%%%%%%%%%%%%%%%%%%%%%%%%%%%%%%%%%%%%%%%%%%%%%%%%%%%%%%%%%%%
%\eat{
\clearpage
\onecolumn
\appendix

%\begin{center}
%	{\large \bf PAGE: A Simple and Optimal Probabilistic Gradient Estimator for\\ Nonconvex Optimization (Supplementary Material)}
%\end{center}

\section{Missing Proofs for Nonconvex Finite-Sum Problems}
\label{app:finite}
Appendix \ref{app:finite} and Appendix \ref{app:online} provide proof details for nonconvex finite-sum and online problems, respectively.
For the PL setting where faster linear convergence rates can be obtained, Appendix \ref{app:finite-pl} and Appendix \ref{app:online-pl} provide proof details for nonconvex finite-sum and online problems under PL condition, respectively.
Before providing the detailed proofs for main theorems and corollaries, we first provide a lemma of smoothness and a general key technical lemma which are used in the following Appendices \ref{app:finite}--\ref{app:online-pl} regardless of the settings.

\begin{lemma}\label{lem:smooth}
	If function $f(x) :=\frac{1}{n}\sum_{i=1}^{n}f_i(x)$ is average $L$-smooth (see Assumption \ref{asp:smooth}), i.e., if 
	\begin{align}\label{eq:avg-smooth}
	\E_i[\ns{\nabla f_i(x) - \nabla f_i(y)}]\leq L^2 \ns{x-y}, \quad \forall x,y \in \R^d,
	\end{align}
	then $f$ is also $L$-smooth, i.e., $\n{\nabla f(x) - \nabla f(y)} \leq L \n{x-y}$ and thus 
	\begin{align}\label{eq:smooth}
	f(y) \leq f(x) + \inner{\nabla f(x)}{y-x} + \frac{L}{2}\ns{y-x},  \quad \forall x,y \in \R^d.
	\end{align}
\end{lemma}
\begin{proofof}{Lemma \ref{lem:smooth}}
	First, we show the $L$-smoothness of $f$:
	\begin{eqnarray}
	\n{\nabla f(x) - \nabla f(y)} 
	&=& \sqrt{\ns{\E_i[\nabla f_i(x)-\nabla f_i(y)]}}  \notag\\
	&\leq & \sqrt{\E_i[\ns{\nabla f_i(x)-\nabla f_i(y)}]}  \notag \\
	&\overset{\eqref{eq:avg-smooth}}{\leq} &  \sqrt{L^2\ns{x-y}} \notag\\
	&=& L\n{x-y}, \label{eq:L-smooth}
	\end{eqnarray}	
	where the first inequality uses Jensen's inequality: $g(\E[x]) \leq \E[g(x)]$ for a convex function $g$. Then, inequality \eqref{eq:smooth} holds due to standard arguments (we do not claim any novelty here and include the following arguments for completeness):
	\begin{eqnarray}	
	f(y) &=&  f(x) + \int_{0}^{1} \inner{\nabla f(x + \tau (y-x))}{y-x} d\tau \notag\\
	&= & f(x) + \inner{\nabla f(x)}{y-x} + \int_{0}^{1} \inner{\nabla f(x + \tau (y-x))-\nabla f(x)}{y-x} d\tau \notag\\
	&\leq & f(x) + \inner{\nabla f(x)}{y-x} + \int_{0}^{1}\n{\nabla f(x + \tau (y-x))-\nabla f(x)}\n{y-x} d\tau \notag\\
	&\overset{\eqref{eq:L-smooth}}{\leq}  &
	f(x) + \inner{\nabla f(x)}{y-x}  + \int_{0}^{1}L\tau\ns{y-x} d\tau \notag\\
	& =&  f(x) + \inner{\nabla f(x)}{y-x}  + \frac{L}{2}\ns{y-x},
	\end{eqnarray}	
	where the first inequality uses Cauchy–Schwarz inequality $\inner{u}{v} \leq \n{u}\n{v}$.
\end{proofof}

Now, we provide a key Lemma \ref{lem:relation} which describes a useful relation between the function values after and before a gradient descent step, i.e.,  between $f(\xtn)$ and $f(\xt)$ with $\xtn := \xt - \eta \gt$ for any gradient estimator $g^t \in \R^d$ and stepsize $\eta >0$.
\begin{lemma}\label{lem:relation}
	Suppose that function $f$ is $L$-smooth and let $\xtn := \xt - \eta \gt$. Then for any $g^t\in \R^d$ and $\eta>0$, we have
	\begin{align}\label{eq:relation}
	f(\xtn) \leq f(\xt) - \frac{\eta}{2} \ns{\nabla f(\xt)} 
	- \Big(\frac{1}{2\eta} - \frac{L}{2}\Big) \ns{\xtn -\xt}
	+ \frac{\eta}{2}\ns{\gt - \nabla f(\xt)}.
	\end{align}
\end{lemma}
\begin{proofof}{Lemma \ref{lem:relation}}
	Let $\bxtn := \xt -\eta \nabla f(\xt)$. In view of $L$-smoothness of $f$, we have 
	\begin{eqnarray}	
	f(\xtn) 
	&\overset{\eqref{eq:smooth}}{\leq}&
	f(\xt) 
	+ \inner{\nabla f(\xt)}{\xtn-\xt}  
	+ \frac{L}{2}\ns{\xtn-\xt}  \notag \\
	&=& f(\xt)
	+ \inner{\nabla f(\xt)-\gt}{\xtn-\xt}
	+ \inner{\gt}{\xtn-\xt}
	+ \frac{L}{2}\ns{\xtn-\xt} \notag \\
	&=& f(\xt)
	+ \inner{\nabla f(\xt)-\gt}{-\eta \gt}
	- \Big(\frac{1}{\eta}- \frac{L}{2}\Big)\ns{\xtn-\xt} \notag \\
	&=& f(\xt)
	+ \eta\ns{\nabla f(\xt)-\gt}
	- \eta\inner{\nabla f(\xt)-\gt}{\nabla f(\xt)}
	- \Big(\frac{1}{\eta}- \frac{L}{2}\Big)\ns{\xtn-\xt} \notag \\
	&=&f(\xt)
	+ \eta\ns{\nabla f(\xt)-\gt}
	- \frac{1}{\eta}\inner{\xtn-\bxtn}{\xt-\bxtn}
	- \Big(\frac{1}{\eta}- \frac{L}{2}\Big)\ns{\xtn-\xt} \notag \\
	&=& f(\xt)
	+ \eta\ns{\nabla f(\xt)-\gt}
	- \Big(\frac{1}{\eta}- \frac{L}{2}\Big)\ns{\xtn-\xt} \notag\\
	&& \qquad\qquad\qquad -\frac{1}{2\eta}\Big(\ns{\xtn-\bxtn}+\ns{\xt-\bxtn}
	-\ns{\xtn-\xt}\Big) \notag \\
	&=& f(\xt)
	+ \eta\ns{\nabla f(\xt)-\gt}
	- \Big(\frac{1}{\eta}- \frac{L}{2}\Big)\ns{\xtn-\xt} \notag\\
	&& \qquad\qquad\qquad -\frac{1}{2\eta}\Big(\eta^2\ns{\nabla f(\xt)-\gt}+\eta^2\ns{\nabla f(\xt)}
	-\ns{\xtn-\xt}\Big) \notag \\
	&=& f(\xt) 
	- \frac{\eta}{2} \ns{\nabla f(\xt)} 
	- \Big(\frac{1}{2\eta} - \frac{L}{2}\Big) \ns{\xtn -\xt}
	+ \frac{\eta}{2}\ns{\gt - \nabla f(\xt)}. \notag
	\end{eqnarray}	
\end{proofof}

Now, we are ready to provide the detailed proofs for our main convergence theorem and corollaries for \page in the nonconvex finite-sum case (i.e., problem \eqref{prob:finite}).

\subsection{Proof of Main Theorem \ref{thm:finite}}
\label{sec:proof-finite}
In this appendix, we first restate our main convergence result (Theorem~\ref{thm:finite}) in the nonconvex finite-sum case and then provide its proof.

\begingroup
\def\thetheorem{\ref{thm:finite}}
\begin{theorem}[Main theorem for nonconvex finite-sum problem \eqref{prob:finite}]
	Suppose that Assumption \ref{asp:smooth} holds. 
	Choose the stepsize 
	$$\eta \leq \frac{1}{L\left(1+\sqrt{\frac{1-p}{pb'}}\right)},$$
	minibatch size $b=n$, secondary minibatch size $b'<b$, and probability $p_t\equiv p\in (0,1]$.
	Then the number of iterations performed by \page sufficient for  finding an $\epsilon$-approximate solution (i.e., $\E[\n{\nabla f(\hx_T)}]\leq \epsilon$) of nonconvex finite-sum problem \eqref{prob:finite} can be bounded by
	\begin{align}
	\compactify	T = \frac{2\fgap L}{\epsilon^2} \left( 1+ \sqrt{\frac{1-p}{pb'}}\right).
	\end{align}
	Moreover, %according to the gradient estimator of \page (Line \ref{line:grad} of Algorithm \ref{alg:page}), we know that it uses $pb + (1-p)b'$ stochastic gradients for each iteration on the expectation. Thus, 
	the number of stochastic gradient computations (i.e., gradient complexity) is 
	\begin{align}
	\compactify	\#\mathrm{grad} =b+ T\left(pb+(1-p)b'\right) = b+ \frac{2\fgap L}{\epsilon^2} \left( 1+ \sqrt{\frac{1-p}{pb'}}\right)\left(pb+(1-p)b'\right).
	\end{align}
	Note that the first $b$ in $\#\mathrm{grad}$ is due to the computation of $g^0$ (see Line \ref{line:sgd} in Algorithm \ref{alg:page}).
\end{theorem}
\addtocounter{theorem}{-1}
\endgroup

\begin{proofof}{Theorem \ref{thm:finite}}
	Note that since the average $L$-smoothness assumption (Assumption \ref{asp:smooth}) holds for $f$,  we know that $f$ is also $L$-smooth according to Lemma \ref{lem:smooth}.
	Then according to the update step $\xtn := \xt - \eta \gt$ (see Line \ref{line:update} in Algorithm \ref{alg:page}) and Lemma \ref{lem:relation}, we have 
	\begin{align}\label{eq:ft}
	f(\xtn) \leq f(\xt) - \frac{\eta}{2} \ns{\nabla f(\xt)} 
	- \Big(\frac{1}{2\eta} - \frac{L}{2}\Big) \ns{\xtn -\xt}
	+ \frac{\eta}{2}\ns{\gt - \nabla f(\xt)}.
	\end{align}
	
	Now, we use the following Lemma \ref{lem:var-finite} to bound the last variance term of \eqref{eq:ft} for this finite-sum case.
	\begin{lemma}\label{lem:var-finite}
		Suppose that Assumption \ref{asp:smooth} holds. If the gradient estimator $\gtn$ is defined in Line \ref{line:grad} of Algorithm \ref{alg:page}, then we have
		\begin{align}
		\E[\ns{\gtn-\nabla f(\xtn)}] \leq  (1-p_t) \ns{g^{t} - \nabla f(\xt)} + \frac{(1-p_t) L^2}{b'}\ns{\xtn - \xt}. \label{eq:var-finite} 
		\end{align}
	\end{lemma}
	\begin{proofof}{Lemma \ref{lem:var-finite}}
		According to the definition of \page gradient estimator in Line \ref{line:grad} of Algorithm \ref{alg:page}: 
		\begin{align}\label{eq:gt}
		g^{t+1} = \begin{cases}
		\frac{1}{b} \sum \limits_{i\in I} \nabla f_i(x^{t+1}) &\text{with probability } p_t,\\
		g^{t}+\frac{1}{b'} \sum \limits_{i\in I'} (\nabla f_i(x^{t+1})- \nabla f_i(x^{t})) &\text{with probability } 1-p_t.
		\end{cases}
		\end{align}
		A direct calculation now reveals that
		\begin{align}
		&\E[\ns{\gtn-\nabla f(\xtn)}]  \notag\\
		&\overset{\eqref{eq:gt}}{=} p_t \EB{\nsB{\frac{1}{b} \sum_{i\in I} \nabla f_i(x^{t+1}) - \nabla f(\xtn)}} 
		+(1-p_t) \EB{\nsB{g^{t}+\frac{1}{b'} \sum_{i\in I'} (\nabla f_i(x^{t+1})- \nabla f_i(x^{t})) - \nabla f(\xtn)}} \notag\\
		& = (1-p_t) \EB{\nsB{g^{t}+\frac{1}{b'} \sum_{i\in I'} (\nabla f_i(x^{t+1})- \nabla f_i(x^{t})) - \nabla f(\xtn)}}  \label{eq:use-b} \\
		&= (1-p_t) \EB{\nsB{g^{t} - \nabla f(\xt) +\frac{1}{b'} \sum_{i\in I'} (\nabla f_i(x^{t+1})- \nabla f_i(x^{t})) - \nabla f(\xtn)+ \nabla f(\xt) }} \notag\\
		& = (1-p_t) \EB{\nsB{\frac{1}{b'} \sum_{i\in I'} (\nabla f_i(x^{t+1})- \nabla f_i(x^{t})) - \nabla f(\xtn)+ \nabla f(\xt)}}  
		+ (1-p_t) \ns{g^{t} - \nabla f(\xt)}\notag\\
		& = \frac{1-p_t}{b'^2} \EB{\sum_{i\in I'} \nsB{\left(\nabla f_i(x^{t+1})- \nabla f_i(x^{t})\right) - \left(\nabla f(\xtn)- \nabla f(\xt)\right)}} 	+ (1-p_t) \ns{g^{t} - \nabla f(\xt)} \notag\\
		&\leq \frac{1-p_t}{b'} \E[\ns{\nabla f_i(x^{t+1})- \nabla f_i(x^{t})}] + (1-p_t)\ns{g^{t} - \nabla f(\xt)} \notag\\
		&\leq \frac{(1-p_t) L^2}{b'}\ns{\xtn - \xt}  + (1-p_t) \ns{g^{t} - \nabla f(\xt)}, \label{eq:use-averagesmooth}
		\end{align}
		where \eqref{eq:use-b} holds since we let $b=n$ in this finite-sum case, the last inequality \eqref{eq:use-averagesmooth} is due to the average $L$-smoothness Assumption \ref{asp:smooth} (i.e., \eqref{eq:avgsmooth}).
	\end{proofof}
	
	Now, we continue to prove Theorem \ref{thm:finite} using Lemma \ref{lem:var-finite}. We add \eqref{eq:ft} with $\frac{\eta}{2p}$ $\times$ \eqref{eq:var-finite} (here we simply let $p_t \equiv p$), and take expectation to get
	\begin{align}
	&\EB{ f(\xtn) -f^* + \frac{\eta}{2p}\ns{\gtn-\nabla f(\xtn)}}  \notag \\
	&\leq \EB{ f(\xt) -f^* - \frac{\eta}{2} \ns{\nabla f(\xt)} 
		- \Big(\frac{1}{2\eta} - \frac{L}{2}\Big) \ns{\xtn -\xt}
		+ \frac{\eta}{2}\ns{\gt - \nabla f(\xt)}} \newl
	+ \frac{\eta}{2p} \EB{(1-p) \ns{g^{t} - \nabla f(\xt)} + \frac{(1-p) L^2}{b'}\ns{\xtn - \xt}} \notag\\
	&= \EB{f(\xt) -f^* + \frac{\eta}{2p}\ns{\gt-\nabla f(\xt)}  - \frac{\eta}{2} \ns{\nabla f(\xt)} 
		-\Big(\frac{1}{2\eta} - \frac{L}{2} -\frac{(1-p)\eta L^2}{2pb'}\Big) \ns{\xtn -\xt}} \notag\\
	&\leq \EB{f(\xt) -f^* + \frac{\eta}{2p}\ns{\gt-\nabla f(\xt)}  - \frac{\eta}{2} \ns{\nabla f(\xt)}}, \label{eq:use-eta}  
	\end{align}
	where the last inequality \eqref{eq:use-eta} holds due to $\frac{1}{2\eta} - \frac{L}{2} -\frac{(1-p)\eta L^2}{2pb'} \geq 0$ by choosing stepsize 
	\begin{align}\label{eq:eta}
	\eta \leq \frac{1}{L\left(1+\sqrt{\frac{1-p}{pb'}}\right)}.
	\end{align} 
	Now, if we define $\Phi_t := f(\xt) -f^* + \frac{\eta}{2p}\ns{\gt-\nabla f(\xt)}$, then \eqref{eq:use-eta} can be written in the form
	\begin{align}
	\E[\ptn] \leq \E[\pt ] - \frac{\eta}{2} \E[\ns{\nabla f(\xt)}].
	\end{align} 
	Summing up from $t=0$ to $T-1$, we get
	\begin{align}
	\E[\Phi_T] \leq \E[\Phi_0] - \frac{\eta}{2} \sum_{t=0}^{T-1}\E[\ns{\nabla f(\xt)}]. \label{eq:phit}
	\end{align} 
	Then according to the output of \page, i.e.,  $\hx_T$ is randomly chosen from $\{\xt\}_{t\in[T]}$ and $\Phi_0=f(x^0) -f^* + \frac{\eta}{2p}\ns{g^0-\nabla f(x^0)}=f(x^0) -f^* \overset{\text{def}}{=} \fgap$, we have
	\begin{equation}\label{eq:hi890fg8gf8dd}   \E[\ns{\nabla f(\hx_T)}] \leq \frac{2\fgap}{\eta T}.\end{equation}
	If we  set the number of iterations 	 as
	\begin{align}
	T= \frac{2\fgap}{\epsilon^2 \eta}  
	\overset{\eqref{eq:eta}}{=}  \frac{2\fgap L}{\epsilon^2}  \left(1+\sqrt{\frac{1-p}{pb'}}\right),
	\end{align}
	then \eqref{eq:hi890fg8gf8dd}  and Jensen's inequality imply
	\[
	\E[\n{\nabla f(\hx_T)}] \leq \sqrt{\E[\ns{\nabla f(\hx_T)}]} \leq \sqrt{\frac{2\fgap}{\eta T} } =\epsilon.
	\]
	
\end{proofof}

\subsection{Proofs of Corollaries \ref{cor:gd} and \ref{cor:finite}}
\label{sec:proofcor-finite}
Similarly, we first restate the corollaries and then provide their proofs respectively.

\begingroup
\def\thecorollary{\ref{cor:gd}}
\begin{corollary}[We recover GD by letting $p_t\equiv 1$]
	Suppose that Assumption \ref{asp:smooth} holds. 
	Choose the stepsize 
	$\eta \leq \frac{1}{L}$,
	minibatch size $b=n$ and probability $p_t\equiv 1$.
	Then \page reduces to GD, and the number of iterations performed by \page  to find an $\epsilon$-approximate solution of the nonconvex finite-sum problem \eqref{prob:finite} can be bounded by
	$
	T = \frac{2\fgap L}{\epsilon^2}.
	$
	Moreover, the number of stochastic gradient computations (i.e., gradient complexity) is 
	\begin{align}
	\compactify	\#\mathrm{grad} = n+ \frac{2\fgap L n}{\epsilon^2} =O\left(\frac{n}{\epsilon^2}\right).
	\end{align}
\end{corollary} 
\addtocounter{corollary}{-1}
\endgroup

\begin{proofof}{Corollary \ref{cor:gd}}
	If the probability is set to $p=1$, the term $\sqrt{\frac{1-p}{pb'}}$  disappears from the stepsize $\eta$, and the total number of iterations $T$ in Theorem~\ref{thm:finite}. So, the bound on the stepsize simplified to  $\eta\leq \frac{1}{L}$, and the total number of iterations simplifies to $T=\frac{2\fgap L}{\epsilon^2}$. We know that the gradient estimator of \page (Line \ref{line:grad} of Algorithm \ref{alg:page}) uses $pb + (1-p)b' = b$ stochastic gradients in each iteration. Thus,	the gradient complexity is $\#\mathrm{grad} = b+Tb = n+ \frac{2\fgap L n}{\epsilon^2}$, as claimed.
\end{proofof}

\begingroup
\def\thecorollary{\ref{cor:finite}}
\begin{corollary}[Optimal result for nonconvex finite-sum problem \eqref{prob:finite}]
	Suppose that Assumption \ref{asp:smooth}  holds. 
	Choose the stepsize 
	$\eta \leq \frac{1}{L(1+\sqrt{b}/b')}$,
	minibatch size $b=n$, secondary minibatch size $b'\leq \sqrt{b}$ and probability $p_t\equiv \frac{b'}{b+b'}$.
	Then the number of iterations performed by \page  to find an $\epsilon$-approximate solution of the nonconvex finite-sum problem \eqref{prob:finite} can be bounded by
	%	\begin{align}
	$T = \frac{2\fgap L}{\epsilon^2} ( 1+ \frac{\sqrt{b}}{b'}).$
	%	\end{align}
	Moreover, the number of stochastic gradient computations (i.e., gradient complexity) is 
	\begin{align} \label{eq:res-finite-app}
	\compactify \#\mathrm{grad} %= b+ T\left(pb+(1-p)b'\right) %\leq  b+ \frac{4\fgap L}{\epsilon^2} \left( b'+ \sqrt{b}\right) 
	\leq n + \frac{8\fgap L \sqrt{n}}{\epsilon^2}  
	= O\left(n+ \frac{\sqrt{n}}{\epsilon^2}\right).
	\end{align}
\end{corollary} 
\addtocounter{corollary}{-1}
\endgroup

\begin{proofof}{Corollary \ref{cor:finite}}
	If we choose probability $p=\frac{b'}{b+b'}$, then  $\sqrt{\frac{1-p}{pb'}}=\frac{\sqrt{b}}{b'}$. 	Thus, according to Theorem \ref{thm:finite}, the stepsize bound becomes $\eta \leq \frac{1}{L(1+\sqrt{b}/b')}$ and the total number of iterations becomes $T=\frac{2\fgap L}{\epsilon^2}( 1+ \frac{\sqrt{b}}{b'})$. We know that the gradient estimator of \page (Line \ref{line:grad} of Algorithm \ref{alg:page}) uses $pb + (1-p)b' = \frac{2bb'}{b+b'}$ stochastic gradients in each iteration  on  expectation. Thus, the gradient complexity  is
	\begin{align*}
	\#\mathrm{grad}  
	&= b+T\left(pb+(1-p)b'\right)  \\
	&= b+ \frac{2\fgap L}{\epsilon^2}\left( 1+ \frac{\sqrt{b}}{b'}\right) \frac{2bb'}{b+b'} \\
	&\leq b+ \frac{2\fgap L}{\epsilon^2}\left( 1+ \frac{\sqrt{b}}{b'}\right) 2b' \\
	%	& \leq b + \frac{4\fgap L}{\epsilon^2}( b'+ \sqrt{b}) \\
	&\leq  n + \frac{8\fgap L \sqrt{n}}{\epsilon^2}, 
	\end{align*}
	where the last inequality is due to the parameter setting $b = n$ and $b'\leq \sqrt{b}$.
\end{proofof}

\subsection{Proof of Theorem \ref{thm:lb}}
Before providing the proof for the lower bound theorem, we recall the standard definition of the algorithm class of linear-span first-order algorithms.
\begin{definition}[Linear-span first-order algorithm]\label{def:linear-span}
	Consider a (randomized) algorithm $\cA$ starting with $x^0$ and let $x^t$ be the point obtained at iteration $t\geq 0$. Then $\cA$ is called a linear-span first-order algorithm if 
	\begin{align} \label{eq:linear-span}
	x^t \in \mathrm{Lin}\{x^0, x^1, \ldots, x^{t-1}, \nabla f_{i_0}(x^0), \nabla f_{i_1}(x^1), \ldots, \nabla f_{i_{t-1}}(x^{t-1})\},
	\end{align}
	where Lin denotes the linear span, and $i_j$ denotes the individual function (or multiple functions) chosen by $\cA$ at iteration $j$. 
\end{definition}

We now restate the lower bound result (Theorem \ref{thm:lb}) and then provide its proof.
\begingroup
\def\thetheorem{\ref{thm:lb}}
\begin{theorem}[Lower bound]
	For any $L>0$, $\fgap>0$ and $n>0$, there exists a  large enough dimension $d$ and a function $f:\R^d\to \R$ satisfying Assumption \ref{asp:smooth} in the finite-sum case  such that any linear-span first-order algorithm needs $\Omega(n + \frac{\fgap L \sqrt{n}}{\epsilon^2})$ stochastic gradient computations in order to finding an $\epsilon$-approximate solution, i.e., a point $\hx$ such that $\E\n{\nabla f(\hx)} \leq \epsilon$.
\end{theorem}
\addtocounter{theorem}{-1}
\endgroup

\begin{proofof}{Theorem \ref{thm:lb}}
	Consider the function $f(x)=\frac{1}{n}\sum_{i=1}^{n} f_i(x)$, where 
	\begin{align}
	f_i(x) := c\inner{v_i}{x}+\frac{L}{2}\ns{x} \label{eq:fi}
	\end{align}
	for some constant $c$. 	First, we show that $f:\R^d\to \R$ satisfies Assumption~\ref{asp:smooth} as follows:
	\begin{align*}
	\E_i[\ns{\nabla f_i(x) - \nabla f_i(y)}] 
	&= 	\E_i[\ns{(cv_i+Lx) - (cv_i+Ly)}]    \notag\\
	&= 	\E_i[\ns{L(x-y)}]    \notag\\
	&= L^2 \ns{x-y}.
	\end{align*}
	Without loss of generality, we assume that $x^0=0$. Otherwise one can consider the shifted function $f(x+x^0)$ instead.
	Now, we compute $\fgap$ as follows:
	\begin{align}
	f(x^0) -f^*
	&= f(0) - f(x^*)  \notag\\
	&= 0 - \left( \frac{c}{n}\sum_{i=1}^n \inner{v_i}{x^*} + \frac{L}{2} \ns{x^*}\right) \notag\\
	&= \frac{c^2}{2Ln^2}\bigg\|\sum_{i=1}^n v_i \bigg\|^2   \label{eq:xstar} \\
	&=\fgap, 
	\end{align}
	where the equality \eqref{eq:xstar} is due to $x^* = -\frac{c}{Ln}\sum_{i=1}^n v_i$,
	and the last equality holds by choosing the appropriate parameter $c$.
	Note that we only need to consider the case $\epsilon \leq O(\sqrt{\fgap L})$ since the gradient norm at the initial point $x^0$ already achieves this order, i.e., $\n{\nabla f(x^0)} \leq \sqrt{2\fgap L}$. Indeed, since
	\begin{align}
	f^* &\leq f\left(x^0-\frac{1}{L}\nabla f(x^0)\right) \notag\\
	&\leq f(x^0) + \left\langle \nabla f(x^0), -\frac{1}{L}\nabla f(x^0) \right\rangle + \frac{L}{2}\left\|\frac{1}{L}\nabla f(x^0) \right\|^2 \label{eq:smoothx0}\\
	&=f(x^0) -\frac{1}{2L} \ns{\nabla f(x^0)}, \notag 
	\end{align}
	where the inequality \eqref{eq:smoothx0} uses the $L$-smoothness of $f$ (see Lemma \ref{lem:smooth}), we have $\n{\nabla f(x^0)} \leq \sqrt{2L(f(x^0)-f^*)} = \sqrt{2\fgap L}$.
	
	Now according to the definition of linear-span first-order algorithms (i.e., Definition \ref{def:linear-span}) and noting that the stochastic gradient is $\nabla f_i(x) = cv_i+Lx$ and $x^0=0$, after querying $t$ stochastic gradients, we have
	\begin{align} \label{eq:xt}
	x^t \in \mathrm{Lin}\{v_{i_0}, v_{i_1},\ldots,v_{i_{t-1}}\},
	\end{align}
	where $i_{0},i_{1},\ldots, i_{t-1}$ denote the $t$ functions which are queried for stochastic gradient computations.
	For the gradient norm, we have 
	\begin{align}\label{eq:gnorm}
	\n{\nabla f(x)} = \bigg\| \frac{c}{n}\sum_{i=1}^{n}v_i + Lx  \bigg\|.
	\end{align}
	If we choose $\{v_i\}_{i\in[n]}$ to be orthogonal vectors, for example, choose $v_1=(1,1,\ldots,1,0,\ldots,0)^T$ (the first $\frac{d}{n}$ elements are 1 and all remaining are 0), $v_2=(0,0,\ldots,0,1,1,\ldots, 1, 0,\ldots,0)^T$ (the elements with indices from $\frac{d}{n} +1$  to $\frac{2d}{n}$ are 1 and others are 0), $\ldots$, $v_i$ (the elements with indices from $\frac{(i-1)d}{n} +1$ to $\frac{id}{n}$ are 1 and others are 0). In other words, we divide the indices $\{1,2,\ldots, d\}$ into $n$ parts, and set one part to be 1 and other parts to be 0 for each $v_i$. Note that $v_i\in \R^d$, for all $i\in [n]$.
	Thus, if fewer than $\frac{n}{2}$ functions have been queried for stochastic gradient computations, then according to \eqref{eq:xt} we know that the current point $x$ belongs to a subspace with dimension at most $\frac{d}{n}\times \frac{n}{2}=\frac{d}{2}$ in $\R^d$. 
	Moreover, according to \eqref{eq:gnorm} we have
	\begin{align}
	\n{\nabla f(x)} \geq  \frac{c}{n} \sqrt{\frac{d}{2}} = \Omega(\epsilon),
	\end{align}
	where the last equality holds by choosing appropriate parameters $c$ and $d$. 
	
	So far, we have shown a lower bound of $\Omega(n)$ stochastic gradient computations for any linear-span first-order algorithm finding an $\epsilon$-approximate solution.
	For the second term $\Omega(\frac{\fgap L \sqrt{n}}{\epsilon^2})$, we directly use the previous lower bound provided by 
	\citet{fang2018spider}. They proved this lower bound term in the small $n$ case, i.e., $n\leq O(\frac{\fgap^2L^2}{\epsilon^4})$.
	Here we recall their lower bound theorem.
	\begin{theorem}[\citealp{fang2018spider}]
		For any $L>0$, $\fgap>0$ and $n\leq O(\frac{\fgap^2L^2}{\epsilon^4})$, there exists a  large enough dimension $d$ and a function $f:\R^d\to \R$ satisfying Assumption \ref{asp:smooth} in the finite-sum case  such that any linear-span first-order algorithm needs $\Omega(\frac{\fgap L \sqrt{n}}{\epsilon^2})$ stochastic gradient computations in order to finding an $\epsilon$-approximate solution, i.e., a point $\hx$ such that $\E\n{\nabla f(\hx)} \leq \epsilon$.
	\end{theorem}
	Now, the lower bound $\Omega(n + \frac{\fgap L \sqrt{n}}{\epsilon^2})$ is proved by combining the term $\Omega(\frac{\fgap L \sqrt{n}}{\epsilon^2})$ in the above theorem and $\Omega(n)$ in our previous arguments.
\end{proofof}

\newpage
\section{Missing Proofs for Nonconvex Online Problems}
\label{app:online}

In this appendix, we provide the detailed proofs for our main convergence theorem and its corollaries for \page in the nonconvex online case (i.e., problem \eqref{prob:online}).
Recall that we refer this online problem \eqref{prob:online} as the finite-sum problem \eqref{prob:finite} with large or infinite $n$. Also, we need the bounded variance assumption (Assumption \ref{asp:bv}) in this online case.

\subsection{Proof of Main Theorem \ref{thm:online}}
\label{sec:proof-online}
Similarly to Appendix \ref{sec:proof-finite}, we first restate the main convergence result (Theorem \ref{thm:online}) in the nonconvex online case and then provide its proof.

\begingroup
\def\thetheorem{\ref{thm:online}}
\begin{theorem}[Main theorem for nonconvex online problem \eqref{prob:online}]
	Suppose that Assumptions \ref{asp:bv} and \ref{asp:smooth} hold. 
	Choose the stepsize 
	$$\eta \leq \frac{1}{L\left(1+\sqrt{\frac{1-p}{pb'}}\right)},$$
	minibatch size $b=\min \{\lceil \frac{2\sigma^2}{\epsilon^2} \rceil, n\}$, secondary minibatch size $b'<b$ and probability $p_t \equiv p\in (0,1]$.
	Then the number of iterations performed by \page  to find an $\epsilon$-approximate solution ($\E[\n{\nabla f(\hat{x}_T)}]\leq \epsilon$) of nonconvex online problem \eqref{prob:online} can be bounded by
	\begin{align}
	\compactify T = \frac{4\fgap L}{\epsilon^2} \left( 1+ \sqrt{\frac{1-p}{pb'}}\right) + \frac{1}{p}.
	\end{align}
	Moreover, %according to the gradient estimator of \page (Line \ref{line:grad} of Algorithm \ref{alg:page}), we know that it uses $pb + (1-p)b'$ stochastic gradients for each iteration on the expectation. Thus, 
	the number of stochastic gradient computations (gradient complexity) is 
	\begin{align}
	\compactify	\#\mathrm{grad} =b+ T\left(pb+(1-p)b'\right)  = 2b+ \frac{(1-p)b'}{p} + \frac{4\fgap L}{\epsilon^2} \left( 1+ \sqrt{\frac{1-p}{pb'}}\right)\left(pb+(1-p)b'\right).
	\end{align}
\end{theorem}
\addtocounter{theorem}{-1}
\endgroup

\begin{proofof}{Theorem \ref{thm:online}}
	Similarly, we know that $f$ is also $L$-smooth according to Lemma \ref{lem:smooth}.
	Then according to the update step $\xtn := \xt - \eta \gt$ (see Line \ref{line:update} in Algorithm \ref{alg:page}) and Lemma \ref{lem:relation}, we have 
	\begin{align}\label{eq:ft-online}
	f(\xtn) \leq f(\xt) - \frac{\eta}{2} \ns{\nabla f(\xt)} 
	- \Big(\frac{1}{2\eta} - \frac{L}{2}\Big) \ns{\xtn -\xt}
	+ \frac{\eta}{2}\ns{\gt - \nabla f(\xt)}.
	\end{align}
	Now, we use the following Lemma \ref{lem:var-online} to bound the last variance term of \eqref{eq:ft-online} for this online case.
	\begin{lemma}\label{lem:var-online}
		Suppose that Assumptions \ref{asp:bv} and \ref{asp:smooth} hold. If the gradient estimator $\gtn$ is defined in Line \ref{line:grad} of Algorithm \ref{alg:page}, then we have
		\begin{align}
			\E[\ns{\gtn-\nabla f(\xtn)}] \leq  (1-p_t) \ns{g^{t} - \nabla f(\xt)} + \frac{(1-p_t) L^2}{b'}\ns{\xtn - \xt} + {\bf 1}_{\{b<n\}} \frac{p_t\sigma^2}{b}. \label{eq:var-online} 
		\end{align}
	\end{lemma}
	\begin{proofof}{Lemma \ref{lem:var-online}}
		According to the definition of \page gradient estimator in Line \ref{line:grad} of Algorithm \ref{alg:page}
		\begin{align}\label{eq:gt-online}
		g^{t+1} = \begin{cases}
		\frac{1}{b} \sum \limits_{i\in I} \nabla f_i(x^{t+1}) &\text{with probability } p_t,\\
		g^{t}+\frac{1}{b'} \sum \limits_{i\in I'} (\nabla f_i(x^{t+1})- \nabla f_i(x^{t})) &\text{with probability } 1-p_t,
		\end{cases}
		\end{align}
		we have 
		\begin{align}
		&\E[\ns{\gtn-\nabla f(\xtn)}]  \notag\\
		&\overset{\eqref{eq:gt-online}}{=} p_t \EB{\nsB{\frac{1}{b} \sum_{i\in I} \nabla f_i(x^{t+1}) - \nabla f(\xtn)}} 
		+(1-p_t) \EB{\nsB{g^{t}+\frac{1}{b'} \sum_{i\in I'} (\nabla f_i(x^{t+1})- \nabla f_i(x^{t})) - \nabla f(\xtn)}} \notag\\
		& = {\bf 1}_{\{b<n\}} \frac{p_t\sigma^2}{b} 
		+(1-p_t) \EB{\nsB{g^{t}+\frac{1}{b'} \sum_{i\in I'} (\nabla f_i(x^{t+1})- \nabla f_i(x^{t})) - \nabla f(\xtn)}}  \label{eq:use-b-online} \\
		&= {\bf 1}_{\{b<n\}} \frac{p_t\sigma^2}{b}  
		+ (1-p_t) \EB{\nsB{g^{t} - \nabla f(\xt) +\frac{1}{b'} \sum_{i\in I'} (\nabla f_i(x^{t+1})- \nabla f_i(x^{t})) - \nabla f(\xtn)+ \nabla f(\xt) }} \notag\\
		& =  {\bf 1}_{\{b<n\}} \frac{p_t\sigma^2}{b}  + (1-p_t) \ns{g^{t} - \nabla f(\xt)} 
		+(1-p_t) \EB{\nsB{\frac{1}{b'} \sum_{i\in I'} (\nabla f_i(x^{t+1})- \nabla f_i(x^{t})) - \nabla f(\xtn)+ \nabla f(\xt)}}  \notag\\
		& = {\bf 1}_{\{b<n\}} \frac{p_t\sigma^2}{b}  + (1-p_t) \ns{g^{t} - \nabla f(\xt)} 
		+\frac{1-p_t}{b'^2} \EB{\sum_{i\in I'} \nsB{\left(\nabla f_i(x^{t+1})- \nabla f_i(x^{t})\right) - \left(\nabla f(\xtn)- \nabla f(\xt)\right)}}  \notag\\
		&\leq {\bf 1}_{\{b<n\}} \frac{p_t\sigma^2}{b}  + (1-p_t) \ns{g^{t} - \nabla f(\xt)}  + \frac{1-p_t}{b'} \E[\ns{\nabla f_i(x^{t+1})- \nabla f_i(x^{t})}] \notag\\
		&\leq {\bf 1}_{\{b<n\}} \frac{p_t\sigma^2}{b}  + (1-p_t) \ns{g^{t} - \nabla f(\xt)} + \frac{(1-p_t) L^2}{b'}\ns{\xtn - \xt}, \label{eq:use-averagesmooth-online}
		\end{align}
		where \eqref{eq:use-b-online} is due to Assumption \ref{asp:bv}, i.e., \eqref{eq:bv} (where ${\bf 1_{\{\cdot\}}}$ denotes the indicator function), the last inequality \eqref{eq:use-averagesmooth-online} is due to the average $L$-smoothness Assumption \ref{asp:smooth}, i.e., \eqref{eq:avgsmooth}.
	\end{proofof}
	
	Now, we continue to prove Theorem \ref{thm:online} using Lemma \ref{lem:var-online}. We add \eqref{eq:ft-online} with $\frac{\eta}{2p}$ $\times$ \eqref{eq:var-online} (here we simply let $p_t \equiv p$), and take expectation to get
	\begin{align}
	&\EB{ f(\xtn) -f^* + \frac{\eta}{2p}\ns{\gtn-\nabla f(\xtn)}}  \notag \\
	&\leq \EB{ f(\xt) -f^* - \frac{\eta}{2} \ns{\nabla f(\xt)} 
		- \Big(\frac{1}{2\eta} - \frac{L}{2}\Big) \ns{\xtn -\xt}
		+ \frac{\eta}{2}\ns{\gt - \nabla f(\xt)}} \newl
	+ \frac{\eta}{2p} \EB{(1-p) \ns{g^{t} - \nabla f(\xt)} + \frac{(1-p) L^2}{b'}\ns{\xtn - \xt} + {\bf 1}_{\{b<n\}} \frac{p\sigma^2}{b}} \notag\\
	&= \E\Big[f(\xt) -f^* + \frac{\eta}{2p}\ns{\gt-\nabla f(\xt)}  - \frac{\eta}{2} \ns{\nabla f(\xt)} 
	+ {\bf 1}_{\{b<n\}} \frac{\eta\sigma^2}{2b} \newl
	-\Big(\frac{1}{2\eta} - \frac{L}{2} -\frac{(1-p)\eta L^2}{2pb'}\Big) \ns{\xtn -\xt}\Big] \notag\\
	&\leq \EB{f(\xt) -f^* + \frac{\eta}{2p}\ns{\gt-\nabla f(\xt)}  - \frac{\eta}{2} \ns{\nabla f(\xt)}  
		+ {\bf 1}_{\{b<n\}} \frac{\eta\sigma^2}{2b}}, \label{eq:use-eta-online}  
	\end{align}
	where the last inequality \eqref{eq:use-eta-online} holds due to $\frac{1}{2\eta} - \frac{L}{2} -\frac{(1-p)\eta L^2}{2pb'} \geq 0$ by choosing stepsize 
	\begin{align}\label{eq:eta-online}
	\eta \leq \frac{1}{L\left(1+\sqrt{\frac{1-p}{pb'}}\right)}.
	\end{align} 
	Now, if we define $\Phi_t := f(\xt) -f^* + \frac{\eta}{2p}\ns{\gt-\nabla f(\xt)}$, then \eqref{eq:use-eta-online} turns to
	\begin{align}
	\E[\ptn] \leq \E[\pt ] - \frac{\eta}{2} \E[\ns{\nabla f(\xt)}] + {\bf 1}_{\{b<n\}} \frac{\eta\sigma^2}{2b}.
	\end{align} 
	Summing up it from $t=0$ for $T-1$, we have 
	\begin{align}
	\E[\Phi_T] \leq \E[\Phi_0] - \frac{\eta}{2} \sum_{t=0}^{T-1}\E[\ns{\nabla f(\xt)}]+{\bf 1}_{\{b<n\}} \frac{\eta T \sigma^2}{2b}.
	\end{align} 
	Then, according to the output of \page, i.e.,  $\hx_T$ is randomly chosen from $\{\xt\}_{t\in[T]}$,
	we have
	\begin{align}\label{eq:last1}
	\E[\ns{\nabla f(\hx_T)}] &\leq \frac{2\E[\Phi_0]}{\eta T} + {\bf 1}_{\{b<n\}} \frac{\sigma^2}{b}.
	\end{align}
	For the term $\E[\Phi_0]$, we have 
	\begin{align}
	\E[\Phi_0] 
	&:=\EB{ f(x^0) -f^* + \frac{\eta}{2p}\ns{g^0-\nabla f(x^0)}} \notag\\
	&=\EB{ f(x^0) -f^*  + \frac{\eta}{2p}\nsB{\frac{1}{b} \sum_{i\in I} \nabla f_i(x^0) -\nabla f(x^0)}} \label{eq:use-g0} \\
	&\leq  f(x^0) -f^*  + {\bf 1}_{\{b<n\}} \frac{\eta \sigma^2}{2pb}, \label{eq:use-b0} 
	\end{align}
	where \eqref{eq:use-g0} follows from the definition of $g^0$ (see Line \ref{line:sgd} of Algorithm \ref{alg:page}), and 
	\eqref{eq:use-b0} is due to Assumption~\ref{asp:bv}, i.e., \eqref{eq:bv} (where ${\bf 1_{\{\cdot\}}}$ denotes the indicator function).	Plugging \eqref{eq:use-b0} into \eqref{eq:last1} and noting that $\fgap:= f(x^0) -f^*$, we have 
	\begin{align}
	\E[\ns{\nabla f(\hx_T)}] &\leq \frac{2\fgap}{\eta T} +  {\bf 1}_{\{b<n\}}  \frac{\sigma^2}{pbT}  + {\bf 1}_{\{b<n\}} \frac{\sigma^2}{b} \notag\\
	&\leq \frac{2\fgap}{\eta T} +  \frac{\epsilon^2}{2pT}  + \frac{\epsilon^2}{2} \label{eq:chooseb}\\
	& =\epsilon^2,\label{eq:last-online}
	\end{align}
	where \eqref{eq:chooseb} follows from the parameter setting of minibatch size $b=\min\{\lceil \frac{2\sigma^2}{\epsilon^2} \rceil, n\}$, and the last equality \eqref{eq:last-online} holds by letting the number of iterations 
	\begin{align}
	T= \frac{4\fgap}{\epsilon^2 \eta} + \frac{1}{p}   
	\overset{\eqref{eq:eta-online}}{=}  \frac{4\fgap L}{\epsilon^2}  \left(1+\sqrt{\frac{1-p}{pb'}}\right) + \frac{1}{p}.
	\end{align}
	
	Now, the proof is finished since
	\begin{align}
	\E[\n{\nabla f(\hx_T)}] \leq \sqrt{\E[\ns{\nabla f(\hx_T)}]} & =\epsilon.
	\end{align}
\end{proofof}

\subsection{Proofs of Corollaries \ref{cor:sgd}, \ref{cor:online} and \ref{cor:lbonline}}
\label{sec:proofcor-online}
Similarly to Appendix \ref{sec:proofcor-finite}, we first restate the corollaries in this online case and then provide their proofs, respectively.

\begingroup
\def\thecorollary{\ref{cor:sgd}}
\begin{corollary}[We recover SGD by letting $p_t \equiv 1$]
	Suppose that Assumptions \ref{asp:bv} and \ref{asp:smooth} hold. 
	Let stepsize 
	$\eta \leq \frac{1}{L}$,
	minibatch size $b=\lceil \frac{2\sigma^2}{\epsilon^2} \rceil$ and probability $p_t \equiv 1$,
	then the number of iterations performed by \page to find an $\epsilon$-approximate solution of nonconvex online problem \eqref{prob:online}  can be bounded by
	$
	T = \frac{4\fgap L}{\epsilon^2} +1.
	$ 
	Moreover, the number of stochastic gradient computations (gradient complexity) is 
	\begin{align}
	\compactify	\#\mathrm{grad} = \frac{4\sigma^2}{\epsilon^2}+ \frac{8\fgap L \sigma^2}{\epsilon^4}  = O\left(\frac{\sigma^2}{\epsilon^4}\right).
	\end{align}
\end{corollary} 
\addtocounter{corollary}{-1}
\endgroup

\begin{proofof}{Corollary \ref{cor:sgd}}
	If the probability parameter is set to $p=1$, then $\sqrt{\frac{1-p}{pb'}}$  disappears from the stepsize $\eta$, and the total number of iterations $T$ in Theorem~\ref{thm:online}. Hence, the stepsize rule simplifies to $\eta\leq \frac{1}{L}$, and the total number of iterations becomes $T=\frac{4\fgap L}{\epsilon^2} +1$. We know that the gradient estimator of \page (Line \ref{line:grad}  uses $pb + (1-p)b' = b$ stochastic gradients in each iteration. Thus,	the gradient complexity is $\#\mathrm{grad} = b+Tb = \frac{4\sigma^2}{\epsilon^2}+ \frac{8\fgap L \sigma^2}{\epsilon^4}$.
\end{proofof}

\begingroup
\def\thecorollary{\ref{cor:online}}
\begin{corollary}[Optimal result for nonconvex online problem \eqref{prob:online}]
	Suppose that Assumptions \ref{asp:bv} and \ref{asp:smooth} hold. 
	Choose the stepsize 
	$\eta \leq \frac{1}{L(1+\sqrt{b}/b')}$,
	minibatch size $b=\min \{\lceil \frac{2\sigma^2}{\epsilon^2} \rceil, n\}$, secondary minibatch size $b'\leq \sqrt{b}$ and probability $p_t \equiv \frac{b'}{b+b'}$.
	Then the number of iterations performed by \page sufficient to find an $\epsilon$-approximate solution of nonconvex online problem \eqref{prob:online} can be bounded by
	%	\begin{align}
	$T = \frac{4\fgap L}{\epsilon^2} ( 1+ \frac{\sqrt{b}}{b'}) +\frac{b+b'}{b'}.$
	%	\end{align}
	Moreover, 
	%	according to the gradient estimator of \page (Line \ref{line:grad} of Algorithm \ref{alg:page}), we know that it uses $pb + (1-p)b'=\frac{2bb'}{b+b'}<2b'$ stochastic gradients for each iteration on the expectation. Thus, 
	the number of stochastic gradient computations (i.e., gradient complexity) is 
	\begin{align}
	\compactify	\#\mathrm{grad} %= b+ T\left(pb+(1-p)b'\right) %\leq  b+ \frac{8\fgap L}{\epsilon^2} \left( b'+ \sqrt{b}\right) 
	\leq 3b+ \frac{16\fgap L \sqrt{b}}{\epsilon^2}
	=O\left(b + \frac{\sqrt{b}}{\epsilon^2}\right).
	\end{align}
\end{corollary} 
\addtocounter{corollary}{-1}
\endgroup

\begin{proofof}{Corollary \ref{cor:online}}
	If we choose probability $p=\frac{b'}{b+b'}$, then  $\sqrt{\frac{1-p}{pb'}}=\frac{\sqrt{b}}{b'}$. 
	Thus, according to Theorem \ref{thm:online}, the stepsize bound becomes $\eta \leq \frac{1}{L(1+\sqrt{b}/b')}$ and the total number of iterations becomes $T=\frac{4\fgap L}{\epsilon^2}( 1+ \frac{\sqrt{b}}{b'}) + \frac{b+b'}{b'}$. Since the gradient estimator of \page (Line \ref{line:grad} of Algorithm \ref{alg:page}) uses $pb + (1-p)b' = \frac{2bb'}{b+b'}$ stochastic gradients in each iteration in expectation,   the gradient complexity is
	\begin{align*}
	\#\mathrm{grad}  
	&= b+T\left(pb+(1-p)b'\right)  \\
	&= b+ \left(\frac{4\fgap L}{\epsilon^2}\Big( 1+ \frac{\sqrt{b}}{b'}\Big) + \frac{b+b'}{b'}\right)\frac{2bb'}{b+b'} \\
	&= 3b+ \frac{4\fgap L}{\epsilon^2}\Big( 1+ \frac{\sqrt{b}}{b'}\Big)\frac{2bb'}{b+b'} \\
	&\leq  3b+ \frac{4\fgap L}{\epsilon^2}\Big( 1+ \frac{\sqrt{b}}{b'}\Big)2b'\\
	%	& = 3b + \frac{8\fgap L}{\epsilon^2}( b'+ \sqrt{b}) \\
	&\leq  3b + \frac{16\fgap L \sqrt{b}}{\epsilon^2},
	\end{align*}
	where the last inequality is due to the parameter setting $b'\leq \sqrt{b}$.
\end{proofof}

\begingroup
\def\thecorollary{\ref{cor:lbonline}}
\begin{corollary}[Lower bound]
	For any $L>0$, $\fgap>0$, $\sigma^2>0$ and $n>0$, there exists a large enough dimension $d$ and a function $f:\R^d\to \R$ satisfying Assumptions \ref{asp:bv} and \ref{asp:smooth} in the online case (here $n$ may be finite) such that any linear-span first-order algorithm needs $\Omega(b + \frac{\fgap L \sqrt{b}}{\epsilon^2})$, where $b = \min\{\frac{\sigma^2}{\epsilon^2}, n\}$,  stochastic gradient computations for finding an $\epsilon$-approximate solution, i.e., a point $\hx$ such that $\E\n{\nabla f(\hx)} \leq \epsilon$.
\end{corollary}
\addtocounter{corollary}{-1}
\endgroup
\begin{proofof}{Corollary \ref{cor:lbonline}}
	This lower bound directly follows from the lower bound Theorem \ref{thm:lbonline} given by \citet{arjevani2019lower} and our Theorem \ref{thm:lb}.
\end{proofof}

\newpage
\section{Missing Proofs for Nonconvex Finite-Sum Problems under PL Condition}
\label{app:finite-pl}
In this appendix, we provide detailed proofs for the main convergence theorem and its corollary for nonconvex finite-sum problems under the PL condition (i.e., Assumption \ref{asp:pl}). 

Similar to Lemma \ref{lem:relation}, we provide the following Lemma \ref{lem:relation-pl} which describes a useful relation between the function values after and before a gradient descent step in this PL setting.

\begin{lemma}\label{lem:relation-pl}
	Suppose that function $f$ is $L$-smooth and satisfies PL condition \eqref{eq:pl}. Let $\xtn := \xt - \eta \gt$. Then for any $g^t\in \R^d$ and $\eta>0$, we have
	\begin{align}\label{eq:relation-pl}
	f(\xtn) -f^* \leq (1-\mu \eta) (f(\xt) -f^* )
	- \Big(\frac{1}{2\eta} - \frac{L}{2}\Big) \ns{\xtn -\xt}
	+ \frac{\eta}{2}\ns{\gt - \nabla f(\xt)}.
	\end{align}
\end{lemma}
\begin{proofof}{Lemma \ref{lem:relation-pl}}
	According to Lemma \ref{lem:relation}, we have 
	\begin{align}\label{eq:ft-pl}
	f(\xtn) \leq f(\xt) - \frac{\eta}{2} \ns{\nabla f(\xt)} 
	- \Big(\frac{1}{2\eta} - \frac{L}{2}\Big) \ns{\xtn -\xt}
	+ \frac{\eta}{2}\ns{\gt - \nabla f(\xt)}.
	\end{align}
	Then, by plugging the PL condition \eqref{eq:pl}, i.e., 
	$$\ns{\nabla f(x)} \geq 2\mu (f(x)-f^*),$$ 
	into \eqref{eq:ft-pl}, we get 
	\begin{align}\label{eq:ft-pl2}
	f(\xtn) -f^* \leq (1-\mu \eta) (f(\xt) -f^* )
	- \Big(\frac{1}{2\eta} - \frac{L}{2}\Big) \ns{\xtn -\xt}
	+ \frac{\eta}{2}\ns{\gt - \nabla f(\xt)}.
	\end{align}
\end{proofof}

Now we restate the main convergence theorem under the PL condition and then provide its proof.
\begingroup
\def\thetheorem{\ref{thm:finite-pl}}
\begin{theorem}[Main theorem for nonconvex finite-sum problem \eqref{prob:finite} under PL condition]
	Suppose that Assumptions \ref{asp:smooth} and \ref{asp:pl} hold. 
	Choose the stepsize 
	$$\eta \leq \min \left\{ \frac{1}{L\left(1+\sqrt{\frac{1-p}{pb'}}\right)},~  \frac{p}{2\mu}  \right\},$$
	minibatch size $b=n$, secondary minibatch size $b'<b$, and probability $p_t \equiv  p\in (0,1]$.
	Then the number of iterations performed by \page sufficient for  finding an $\epsilon$-solution ($\E[f(x^T)-f^*]\leq \epsilon$) of nonconvex finite-sum problem \eqref{prob:finite} can be bounded by
	\begin{align}
	\compactify	T =  \left(\left(1+\sqrt{\frac{1-p}{pb'}}\right) \kappa + \frac{2}{p} \right) \log \frac{\fgap}{\epsilon}.
	\end{align}
	Moreover, %according to the gradient estimator of \page (Line \ref{line:grad} of Algorithm \ref{alg:page}), we know that it uses $pb + (1-p)b'$ stochastic gradients for each iteration on the expectation. Thus, 
	the number of stochastic gradient computations (i.e., gradient complexity) is 
	\begin{align}
	\compactify	\#\mathrm{grad} =b+ T\left(pb+(1-p)b'\right) = b+  \left(pb+(1-p)b'\right) \left(\left(1+\sqrt{\frac{1-p}{pb'}}\right) \kappa + \frac{2}{p} \right)\log \frac{\fgap}{\epsilon}.
	\end{align}
\end{theorem}
\addtocounter{theorem}{-1}
\endgroup

\begin{proofof}{Theorem \ref{thm:finite-pl}}
	According to Lemma \ref{lem:relation-pl} and Lemma \ref{lem:var-finite}, we add \eqref{eq:relation-pl} with $\beta$ $\times$ \eqref{eq:var-finite} (here we simply let $p_t \equiv p$), and take expectation to get
	\begin{align}
	&\EB{ f(\xtn) -f^* + \beta\ns{\gtn-\nabla f(\xtn)}}  \notag \\
	&\leq \EB{ (1-\mu \eta)(f(\xt) -f^*) 
		- \Big(\frac{1}{2\eta} - \frac{L}{2}\Big) \ns{\xtn -\xt}
		+ \frac{\eta}{2}\ns{\gt - \nabla f(\xt)}} \newll
	+ \beta \EB{(1-p) \ns{g^{t} - \nabla f(\xt)} + \frac{(1-p) L^2}{b'}\ns{\xtn - \xt}} \notag\\
	&= \E\bigg[(1-\mu\eta)(f(\xt) -f^*) + \left(\frac{\eta}{2}+(1-p)\beta\right)\ns{\gt-\nabla f(\xt)} 
	\newll
	-\Big(\frac{1}{2\eta} - \frac{L}{2} -\frac{(1-p)\beta L^2}{b'}\Big) \ns{\xtn -\xt}\bigg]\notag\\
	&\leq \EB{(1-\mu\eta)\left(f(\xt) -f^* + \beta\ns{\gt-\nabla f(\xt)} \right)}, \label{eq:use-eta-pl}  
	\end{align}
	where the last inequality \eqref{eq:use-eta-pl} holds by choosing the stepsize
	\begin{align}\label{eq:eta-pl}
	\eta \leq \min \left\{ \frac{1}{L\left(1+\sqrt{\frac{1-p}{pb'}}\right)},~  \frac{p}{2\mu}  \right\},
	\end{align}
	and $\beta\geq \frac{\eta}{p}$.
	Now, we define $\Phi_t := f(\xt) -f^* + \beta \ns{\gt-\nabla f(\xt)}$, then \eqref{eq:use-eta-pl} turns to
	\begin{align}
	\E[\ptn] \leq (1-\mu\eta)\E[\pt ].
	\end{align} 
	Telescoping it from $t=0$ for $T-1$, we have 
	\begin{align}
	\E[\Phi_T] \leq (1-\mu\eta)^T\E[\Phi_0].
	\end{align} 
	Note that $\Phi_0=f(x^0) -f^* + \beta\ns{g^0-\nabla f(x^0)}=f(x^0) -f^* \overset{\text{def}}{=} \fgap$, we have
	\begin{align}
	\E[f(x^T)-f^*]\leq (1-\mu\eta)^T \fgap =\epsilon, \label{eq:last-pl}
	\end{align}
	where the last equality \eqref{eq:last-pl} holds by letting the number of iterations 
	\begin{align}
	T=\frac{1}{\mu\eta}\log \frac{\fgap}{\epsilon}
	\overset{\eqref{eq:eta-pl}}{=}\left(\left(1+\sqrt{\frac{1-p}{pb'}}\right) \kappa + \frac{2}{p} \right)\log \frac{\fgap}{\epsilon},
	\end{align}
	where $\kappa:=\frac{L}{\mu}$.
\end{proofof}

Now, we restate the its corollary in which a detailed convergence result is obtained by giving a specific parameter setting and then provide its proof.
\begingroup
\def\thecorollary{\ref{cor:finite-pl}}
\begin{corollary}[Nonconvex finite-sum problem \eqref{prob:finite} under PL condition]
	Suppose that Assumptions \ref{asp:smooth} and \ref{asp:pl} hold. 
	Let stepsize $\eta \leq \min \{ \frac{1}{L(1+\sqrt{b}/b')},~  \frac{b'}{2\mu (b+b')}  \}$,
	minibatch size $b=n$, secondary minibatch size $b'\leq \sqrt{b}$, and probability $p_t\equiv \frac{b'}{b+b'}$.
	Then the number of iterations performed by \page to find an $\epsilon$-solution of nonconvex finite-sum problem \eqref{prob:finite} can be bounded by
	%	\begin{align}
	$T = \left((1+\frac{\sqrt{b}}{b'}) \kappa + \frac{2(b+b')}{b'} \right)  \log \frac{\fgap}{\epsilon}.$
	%	\end{align}
	Moreover, the number of stochastic gradient computations (gradient complexity) is 
	\begin{align} \label{eq:res-finite-pl-app}
	\compactify \#\mathrm{grad} %= b+ T\left(pb+(1-p)b'\right) 
	\leq n+ (4\sqrt{n}\kappa + 4n) \log \frac{\fgap}{\epsilon}
	= O\left((n + \sqrt{n}\kappa)\log \frac{1}{\epsilon} \right).
	\end{align}
\end{corollary} 
\addtocounter{corollary}{-1}
\endgroup

\begin{proofof}{Corollary \ref{cor:finite-pl}}
	If we choose probability $p=\frac{b'}{b+b'}$, then this term $\sqrt{\frac{1-p}{pb'}}=\frac{\sqrt{b}}{b'}$. 
	Thus, according to Theorem \ref{thm:finite-pl}, the stepsize $\eta \leq \min \{ \frac{1}{L(1+\sqrt{b}/b')},~   \frac{b'}{2\mu (b+b')}  \}$ and the total number of iterations $T=\left((1+\frac{\sqrt{b}}{b'}) \kappa + \frac{2(b+b')}{b'} \right)  \log \frac{\fgap}{\epsilon}$. According to the gradient estimator of \page (Line \ref{line:grad} of Algorithm \ref{alg:page}), we know that it uses $pb + (1-p)b' = \frac{2bb'}{b+b'}$ stochastic gradients for each iteration  on the expectation. Thus, the gradient complexity 
	\begin{align*}
	\#\mathrm{grad}  
	&= b+T\left(pb+(1-p)b'\right)  \\
	&= b+ \frac{2bb'}{b+b'} \left((1+\frac{\sqrt{b}}{b'}) \kappa + \frac{2(b+b')}{b'} \right)  \log \frac{\fgap}{\epsilon} \\
	&= b+\left( \frac{2bb'}{b+b'} (1+\frac{\sqrt{b}}{b'}) \kappa +4b \right)  \log \frac{\fgap}{\epsilon}  \\
	& \leq b+\left( 2b'(1+\frac{\sqrt{b}}{b'}) \kappa +4b \right)  \log \frac{\fgap}{\epsilon} \\
	&\leq  n+ (4\sqrt{n}\kappa + 4n) \log \frac{\fgap}{\epsilon},
	\end{align*}
	where the last inequality is due to the parameter setting $b = n$ and $b'\leq \sqrt{b}$.
\end{proofof}

\newpage
\section{Missing Proofs for Nonconvex Online Problems under PL Condition}
\label{app:online-pl}
In this appendix, we provide detailed proofs for the main convergence theorem and its corollary for nonconvex online problems under the PL condition (i.e., Assumption \ref{asp:pl}). 
Recall that we refer this online problem \eqref{prob:online} as the finite-sum problem \eqref{prob:finite} with large or infinite $n$. Also, we need the bounded variance assumption (Assumption \ref{asp:bv}) in this online case.

We first restate the main convergence theorem under the PL condition and then provide its proof.
\begingroup
\def\thetheorem{\ref{thm:online-pl}}
\begin{theorem}[Main theorem for nonconvex online problem \eqref{prob:online} under PL condition]
	Suppose that Assumptions \ref{asp:bv}, \ref{asp:smooth} and \ref{asp:pl} hold. 
	Choose the stepsize 
	$$\eta \leq \min \left\{ \frac{1}{L\left(1+\sqrt{\frac{1-p}{pb'}}\right)},~  \frac{p}{2\mu}  \right\}$$
	minibatch size $b=\min \{\lceil \frac{2\sigma^2}{\mu \epsilon} \rceil, n\}$, secondary minibatch size $b'<b$, and probability $p_t\equiv p\in (0,1]$.
	Then the number of iterations performed by \page sufficient for  finding an $\epsilon$-solution ($\E[f(x^T)-f^*]\leq \epsilon$) of nonconvex finite-sum problem \eqref{prob:finite} can be bounded by
	\begin{align}
	\compactify	T =  \left(\left(1+\sqrt{\frac{1-p}{pb'}}\right) \kappa + \frac{2}{p} \right) \log \frac{2 \fgap}{\epsilon}.
	\end{align}
	Moreover, %according to the gradient estimator of \page (Line \ref{line:grad} of Algorithm \ref{alg:page}), we know that it uses $pb + (1-p)b'$ stochastic gradients for each iteration on the expectation. Thus, 
	the number of stochastic gradient computations (i.e., gradient complexity) is 
	\begin{align}
	\compactify	\#\mathrm{grad} =b+ T\left(pb+(1-p)b'\right) = b+  \left(pb+(1-p)b'\right)\left(\left(1+\sqrt{\frac{1-p}{pb'}}\right) \kappa + \frac{2}{p} \right) \log \frac{2\fgap}{\epsilon}.
	\end{align}
\end{theorem}
\addtocounter{theorem}{-1}
\endgroup

\begin{proofof}{Theorem \ref{thm:online-pl}}
	According to Lemma \ref{lem:relation-pl} and Lemma \ref{lem:var-online}, we add \eqref{eq:relation-pl} with $\beta$ $\times$ \eqref{eq:var-online} (here we simply let $p_t \equiv p$), and take expectation to get
	\begin{align}
	&\EB{ f(\xtn) -f^* + \beta\ns{\gtn-\nabla f(\xtn)}}  \notag \\
	&\leq \EB{ (1-\mu \eta)(f(\xt) -f^*) 
		- \Big(\frac{1}{2\eta} - \frac{L}{2}\Big) \ns{\xtn -\xt}
		+ \frac{\eta}{2}\ns{\gt - \nabla f(\xt)}} \newll
	+ \beta \EB{(1-p) \ns{g^{t} - \nabla f(\xt)} + \frac{(1-p) L^2}{b'}\ns{\xtn - \xt} +{\bf 1}_{\{b<n\}} \frac{p\sigma^2}{b}} \notag\\
	&= \E\bigg[(1-\mu\eta)(f(\xt) -f^*) + \left(\frac{\eta}{2}+(1-p)\beta\right)\ns{\gt-\nabla f(\xt)}  
	+{\bf 1}_{\{b<n\}} \frac{\beta p\sigma^2}{b} 
	\newll
	-\Big(\frac{1}{2\eta} - \frac{L}{2} -\frac{(1-p)\beta L^2}{b'}\Big) \ns{\xtn -\xt}\bigg]\notag\\
	&\leq \EB{(1-\mu\eta)\left(f(\xt) -f^* + \beta\ns{\gt-\nabla f(\xt)} \right) +{\bf 1}_{\{b<n\}} \frac{\beta p\sigma^2}{b} }, \label{eq:use-eta-pl-online}  
	\end{align}
	where the last inequality \eqref{eq:use-eta-pl-online} holds by choosing the stepsize
	\begin{align}\label{eq:eta-pl-online}
	\eta \leq \min \left\{ \frac{1}{L\left(1+\sqrt{\frac{1-p}{pb'}}\right)},~  \frac{p}{2\mu}  \right\},
	\end{align}
	and $\beta\geq \frac{\eta}{p}$.
	Now, we define $\Phi_t := f(\xt) -f^* + \beta \ns{\gt-\nabla f(\xt)}$ and choose  $\beta= \frac{\eta}{p}$, then \eqref{eq:use-eta-pl-online} turns to
	\begin{align}
	\E[\ptn] \leq (1-\mu\eta)\E[\pt ] +{\bf 1}_{\{b<n\}} \frac{\eta\sigma^2}{b}.
	\end{align} 
	Telescoping it from $t=0$ for $T-1$, we have 
	\begin{align}
	\E[\Phi_T] &\leq (1-\mu\eta)^T\E[\Phi_0] + {\bf 1}_{\{b<n\}} \frac{\sigma^2}{b\mu} = \frac{\epsilon}{2} + \frac{\epsilon}{2} \label{eq:last1-pl}
	\end{align} 
	where the last equality \eqref{eq:last1-pl} holds by letting the minibatch size $b=\min \{\lceil \frac{2\sigma^2}{\mu \epsilon} \rceil, n\}$ 
	and the number of iterations 
	\begin{align}
	T=\frac{1}{\mu\eta}\log \frac{2\fgap}{\epsilon}
	\overset{\eqref{eq:eta-pl-online}}{=}\left(\left(1+\sqrt{\frac{1-p}{pb'}}\right) \kappa + \frac{2}{p} \right)\log \frac{2\fgap}{\epsilon},
	\end{align}
	where $\kappa:=\frac{L}{\mu}$.
\end{proofof}

Now, we restate the its corollary in which a detailed convergence result is obtained by giving a specific parameter setting and then provide its proof.
\begingroup
\def\thecorollary{\ref{cor:online-pl}}
\begin{corollary}[Nonconvex online problem \eqref{prob:online} under PL condition]
	Suppose that Assumptions \ref{asp:bv}, \ref{asp:smooth} and \ref{asp:pl} hold. 
	Choose the stepsize 
	$\eta \leq \min \{ \frac{1}{L(1+\sqrt{b}/b')},~  \frac{b'}{2\mu (b+b')}  \}$,
	minibatch size $b=\min \{\lceil \frac{2\sigma^2}{\mu \epsilon} \rceil, n\}$, secondary minibatch $b'\leq \sqrt{b}$ and probability $p_t\equiv \frac{b'}{b+b'}$.
	Then the number of iterations performed by \page  to find an $\epsilon$-solution of nonconvex online problem  \eqref{prob:online} can be bounded by
	%	\begin{align}
	$T =  \left((1+\frac{\sqrt{b}}{b'}) \kappa + \frac{2(b+b')}{b'} \right)  \log \frac{2\fgap}{\epsilon}.$
	%	\end{align}
	Moreover, the number of stochastic gradient computations (gradient complexity) is
	\begin{align}
	\compactify \#\mathrm{grad} %= b+ T\left(pb+(1-p)b'\right) 
	%\leq b+ 4\sqrt{b}\kappa\log \tfrac{2 \fgap}{\epsilon}
	= O\left((b + \sqrt{b}\kappa)\log \frac{1}{\epsilon} \right).
	\end{align}
\end{corollary} 
\addtocounter{corollary}{-1}
\endgroup

\begin{proofof}{Corollary \ref{cor:online-pl}}
	If we choose probability $p=\frac{b'}{b+b'}$, then this term $\sqrt{\frac{1-p}{pb'}}=\frac{\sqrt{b}}{b'}$. 
	Thus, according to Theorem \ref{thm:online-pl}, the stepsize $\eta \leq \min \{ \frac{1}{L(1+\sqrt{b}/b')},~  \frac{b'}{2\mu (b+b')}  \}$ and the total number of iterations $T=\left((1+\frac{\sqrt{b}}{b'}) \kappa + \frac{2(b+b')}{b'} \right) \log \frac{2 \fgap}{\epsilon}$. According to the gradient estimator of \page (Line \ref{line:grad} of Algorithm \ref{alg:page}), we know that it uses $pb + (1-p)b' = \frac{2bb'}{b+b'}$ stochastic gradients for each iteration  on the expectation. Thus, the gradient complexity 
	\begin{align*}
	\#\mathrm{grad}  
	&= b+T\left(pb+(1-p)b'\right)  \\
	&= b+ \frac{2bb'}{b+b'} \left((1+\frac{\sqrt{b}}{b'}) \kappa + \frac{2(b+b')}{b'} \right)  \log \frac{2\fgap}{\epsilon} \\
	&= b+\left( \frac{2bb'}{b+b'} (1+\frac{\sqrt{b}}{b'}) \kappa +4b \right)  \log \frac{2\fgap}{\epsilon}  \\
	& \leq b+\left( 2b'(1+\frac{\sqrt{b}}{b'}) \kappa +4b \right)  \log \frac{2\fgap}{\epsilon} \\
	&\leq  b+ (4\sqrt{b}\kappa + 4b) \log \frac{2\fgap}{\epsilon},
	\end{align*}
	where the last inequality is due to the parameter setting $b'\leq \sqrt{b}$.
\end{proofof}
%}

\end{document}